\documentclass[10pt, a4paper]{article}
\usepackage{lrec}
\usepackage{graphicx}
\usepackage{tabularx}
\usepackage{multirow}
\usepackage{soul}

\usepackage{epstopdf}
\usepackage[latin1]{inputenc}

\usepackage{hyperref}
\usepackage{xstring}
\usepackage{pdfpages}

\title{The Medical Scribe: Corpus Development and Model Performance Analyses}

\name{ \\
Izhak Shafran,
Nan Du,
Linh Tran,
Amanda Perry,
Lauren Keyes,
Mark Knichel,
Ashley Domin,
Lei Huang, \\
Yuhui Chen, 
Gang Li, 
Mingqiu Wang,
Laurent El Shafey,
Hagen Soltau,
Justin S.~Paul
}

\address{Google Inc.}

\abstract{
There is a growing interest in creating tools to assist in clinical note generation using the audio of provider-patient encounters. Motivated by this goal and with the help of providers and medical scribes, we developed an annotation scheme to extract relevant clinical concepts. We used this annotation scheme to label a corpus of about 6k clinical encounters. This was used to train a state-of-the-art tagging model. We report ontologies, labeling results, model performances, and detailed analyses of the results. Our results show that the entities related to medications can be extracted with a relatively high accuracy of 0.90 F-score, followed by symptoms at 0.72 F-score, and conditions at 0.57 F-score. In our task, we not only identify where the symptoms are mentioned but also map them to canonical forms as they appear in the clinical notes. Of the different types of errors, in about 19-38\% of the cases, we find that the model output was correct, and about 17-32\% of the errors do not impact the clinical note. Taken together, the models developed in this work are more useful than the F-scores reflect, making it a promising approach for practical applications. \\
\newline }

\begin{document}

\maketitleabstract

\section{Introduction}
\label{sec:intro}
Medical providers across the United States are required to document clinical visits in the Electronic Health Records. This need for documentation takes up a disproportionate amount of their time and attention, resulting in provider burnout~\cite{Wachter2018,Xu2018}. One study found that full-time primary care physicians spent about 4.5 hours of an 11-hour workday interacting with the clinical documentation systems, yet were still unable to finish their documentation and had to spend an additional 1.4 hours after normal clinical hours~\cite{Arndt2017}.

Speech and natural language processing are now sufficiently mature that there is considerable interest, both in academia and industry, to investigate how these technologies can be exploited to simplify the task of documentation, and to allow providers to dedicate more time to patients. While domain-specific automatic speech recognition (ASR) systems that allow providers to dictate notes have been around for a while, recent work has begun to address the challenges associated with generating clinical notes directly from speech recordings. This includes inducing topic structure from conversation data, extracting relevant information, and clinical summary generation~\cite{Quiroz2019}. In one recent work, authors outlined an end-to-end system; however, the details were scant without empirical evaluations of their building blocks~\cite{Finley2018a}. One of the simplistic approaches uses a hand crafted finite state machine based grammar to locate clinical entities in the ASR transcripts and map them to canonical clinical terms~\cite{HappePBCB03}. This seems to perform well in a narrowly scoped task. A more ambitious approach mapped ASR transcripts to clinical notes by adopting a machine translation approach~\cite{Finley2018b}. However this performed poorly. To address the difficulty in accessing clinical data, researchers have experimented with synthetic data to develop a system for documenting nurse-initiated telephone conversations for congestive heart failure patients who are undergoing telemonitoring after they have been discharged from the hospital~\cite{Liu2019}. In their task, a question-answer based model achieved an F-score of 0.80. This naturally raises the question of how well state-of-art techniques will perform in helping the broader population of clinicians such as primary care providers.

One might expect that the task of extracting clinical concepts from audio faces challenges similar to the domain of unstructured clinical texts. In that domain, one of the earliest public-domain tasks is the {\it i2b2 relations challenge}, defined on a small corpus of written discharge summaries consisting of 394 reports for training, 477 for test, and 877 for evaluation~\cite{Uzuner2011}. Given the small amount of training data, not surprisingly, a disproportionately large number of teams fielded rule-based systems. Conditional random field-based (CRF) systems~\cite{Sutton2011} however did better even with the limited amount of training data~\cite{Uzuner2010}. Other i2b2/n2c2 challenges focused on coreference resolution~\cite{Uzuner2012}, temporal relation extraction~\cite{Sun2013}, drug event extraction~\cite{Henry2019} on medical records, and extracting family history~\cite{Azab2019}. Even though the text was largely unstructured, they benefited from punctuation and capitalization, section headings and other cues in written domain which are unavailable in audio to the same extent. 

With the goal of creating an automated medical scribe, we broke down the task into modular components, including ASR and speaker diarization which are described elsewhere~\cite{Shafey2019}. In this work, we investigate the task of extracting relevant clinical concepts from transcripts. Our key contributions include: (i) defining three tasks -- the Medications Task, the Symptoms Task, and the Conditions Task along with principles employed in developing the annotation guidelines for them (Section~\ref{sec:corpus}); (ii) measuring the label quality using inter-labeler agreements and refining the quality iteratively (Section~\ref{sec:iteration}),  (iii) evaluating the performance of the state-of-the-art models on these tasks (Section~\ref{sec:satmodel}), and (iv) a comprehensive analysis of the performance of the models including manual error categorization (Section~\ref{sec:evaluations}). The corpus we have created in this work is based on private, proprietary data that cannot be publicly shared. Instead, we are sharing the learnings from our experience that might be useful for the wider community as well as the detailed labeling guidelines as supplementary material in the extended version of this paper on arxiv.org.

\section{Corpus Development}
\label{sec:corpus}
The corpus of labeled conversations for this work was developed in conjunction with providers and medical scribes. In our first attempt, we annotated all the relevant clinical information with a comprehensive ontology that was designed for generating clinical notes using a slot-filling approach. This was found to be an extremely challenging task for the medical scribe labelers; it required a lot of cognitive processing, took a long time to label, and the results still had many discrepancies in quality and inter-labeler agreement. 

\subsection{Annotation Guidelines}
The corpus described in this paper was annotated using certain guiding principles, described below. 

First, the cognitive load on the labelers needs to be reasonably low to ensure high inter-labeler agreement. This consideration resulted in labeling each conversation in multiple passes, where the labelers focused only on a small subset of clinical concepts in each pass. They were instructed to ignore clinical information unrelated to the task at hand.

Second, the annotation of all the clinical information necessary to generate a clinical note was broken down into modular tasks, based on the type of entity, specifically, the symptoms, the medications, and the conditions. In each task, labelers focused on annotating the entities along with their coreferences and their attributes. For example, in the Medications Task, the labelers focused on the medications and their associated frequencies, dosages, quantities, and duration. The relationships between them were marked using undirected links. For simplicity, we ignored attributes of attributes, for example, taking one dosage of a medication in the morning and a different dosage in the evening.

Third, the guidelines were refined using a few iterations of experimentation and feedback to improve inter-labeler agreement before labeling the task. Finally, the ontology was pruned to retain clinical concepts with high inter-labeler agreement and sufficiently high occurrences so they could be modeled with the available data. 

\subsection{Description of the Corpus}
The corpus consists of recordings and transcripts of primary care and internal medicine provider-patient encounters. The recordings were split into train, development and test sets consisting of about 5500, 500 and 500 encounters respectively. For measuring the generalization of the model results, the development and test splits were created in such a way that the providers are mutually exclusive. There were no identifiers associated with patients, thus there may be patient overlap across the sets.

\subsection{Symptoms Task}
\label{sec:Sx}
The Symptoms Task focused on extracting symptoms described in the medical encounter so that they could be documented in the History of Present Illness (HPI), Review of Symptoms (ROS) and other sections of the clinical note. Thus, the ontology was designed to mirror the language and organization of symptoms in the note; the colloquial phrasing that patients use to describe symptoms is tagged with a clinically appropriate entity from the ontology. 

\begin{table}[h]
    \centering
    \begin{tabular}{|l|l|} \hline 
    Organ System & 
    Symptom Entities \\ \hline \hline
    \multirow{3}{*}{Constitutional} &
    Const:Fever \\
    & Const:Chills \\
    & Const:Difficulty Sleeping \\ \hline
    \multirow{3}{*}{Gastrointestinal} & 
    GI:Abdominal Distension \\
    & GI:Abdominal Pain \\
    & GI:Vomiting \\ \hline
    \multirow{3}{*}{Neurologic} &
    Neuro:Headache \\ 
    & Neuro:Dizziness \\
    & Neuro:Seizure \\ \hline
    \end{tabular}
    \caption{An excerpt of the entities in the symptoms task.}
    \label{table:SxEntities}
\end{table}

As shown by an excerpt in Table~\ref{table:SxEntities}, the ontology is organized in a categorical fashion, starting with a coarse-grained {\it organ system} which is  further broken down into finer-grained {\it symptom entity tags}. The benefit of this approach is that the organ system can be used to organize symptoms into the ROS section of the note, even if the granular symptom entity isn't correctly inferred. The full ontology contains 186 symptom entities mapped to 14 organ systems.

The attributes for the Symptoms Task, illustrated in Table~\ref{table:SxAttributes}, were chosen based on the HPI elements in the CMS Evaluation and Management Service Documentation Guidelines~\cite{cms}. After the manual labels were generated, the set was pruned to remove entities with low counts and low inter-labeler agreements as these would be hard to model. This resulted in 88 symptom labels.

\begin{table}[h]
    \centering
    \begin{tabular}{|l|l|} \hline
    Type & Attribute \\ \hline \hline
    \multirow{2}{*}{Status} &
    Experienced \\
    & Not Experienced \\ \hline
    \multirow{2}{*}{Property} &
    Duration \\
    & Location \\
    & Severity/Amount \\
    & Frequency \\ \hline
    \end{tabular}
    \caption{The attributes of symptom entities.}
    \label{table:SxAttributes}
\end{table}

The status was marked by adding a second label to the entity. This choice sidestepped the more difficult task of identifying the words that signal whether a symptom was experienced or not. Unlike written text, the status is not explicitly mentioned. Instead, in a conversation, this is implied from the context, spread over multiple speaker turns and not easily associated with specific words.

\begin{table}[t]
    \centering
    \begin{tabular}{ll}
    PT: & I've been having \ul{stomach issues} around here \\
    & for the last \ul{2 weeks}. It's \ul{bad}. \\
    DR: & Okay, in the \ul{upper abdomen}. What does it feel \\ & like? \\
    PT: & It kind of \ul{comes and goes} and \ul{hurts}. \ul{Sometimes} \\
    & I feel \ul{queasy}. \\
    & \\
    \end{tabular}

    \begin{tabular}{r|l} \hline
    Content Span & Symptom Label \\ \hline \hline
    {\it stomach issues} & GI:Other; Experienced \\
    {\it 2 weeks} & Property:Duration \\
    {\it bad} & Property:Severity/Amount \\
    {\it upper abdomen} & Property:Location \\
    {\it comes and goes} & Property:Frequency \\
    {\it hurts} & GI:Abdominal Pain; Experienced\\
    {\it queasy} & GI:Nausea; Status:Experienced \\
    {\it sometimes} & Property:Frequency \\ \hline
    \end{tabular}
    \caption{An example symptoms task with its labels.}
    \label{table:SxEg}
\end{table}

The example in Table~\ref{table:SxEg} illustrates how a conversation was labeled using the ontology for the symptoms task. There were a few challenges in accurately labeling the relevant content words. The annotators were instructed to assign the most specific label, but there were times when the context does not provide sufficient information. In such cases, they are instructed to assign the coarser system category along with a default of {\it Other}. Following this guideline, ``{\it stomach issues}'' in the above example would be assigned {\it GI:Other}.

In addition, we mapped patients' layman description of a symptom to a normalized form. In the illustrative example, the patient's described as feeling ``{\it queasy}'' and that is mapped to the clinical term {\it Nausea}. One advantage of this mapping is that we avoid an explicit normalization step that is commonly used in the literature.

Lastly, in provider-patient conversations, gestures and words are often used to indicate information about the symptoms, such as using the word ``{\it here}'' to refer to the part of the body where the symptom is experienced. The lack of verbalization of this information makes it difficult for an automated system to fully capture all relevant information. However, when providers clarify the location (``{\it upper abdomen}''), this helped in the ultimate task to automate the completion of the note.

\subsection{Medications Task}
\label{sec:rx}
The ontology for the Medications Task was designed to capture information related to all medications ranging from a patient's history to future prescriptions. While the symptoms could be reduced to a closed set (based on how they appear in the clinical notes), the list of medications is large and continually expanding. Therefore, the medication entity was treated as an open set, and the catch all {\it Drug} label was applied to all direct and indirect references to drugs. This included specific and non-specific references, such as ``{\it Tylenol}'', ``{\it the pain medication}'', and ``{\it the medication}''. The attributes related to the medication entity that were annotated include {\it Prop:Frequency}, {\it Prop:Dosage}, {\it Prop:Mode}, and {\it Prop:Duration}. 

In the illustrative example in Table~\ref{table:RxEg}, the entities -- ``{\it diabetes medications}'', ``{\it Sulfonylurea}'', ``{\it Amaryl}'', ``{\it glimepiride}'' -- are annotated with the {\it Drug} label, while ``{\it 1 mg}'', ``{\it pill}'', ``{\it everyday}'' are their attributes.

\begin{table}[t]
    \centering

    \begin{tabular}{ll}
    DR: & Are you taking any \ul{diabetes medication}? \\
    PT: & My kidney doc just changed \ul{the pill}. \\
    DR: &  Oh, a \ul{Sulfonylurea}. Like \ul{Amaryl}? \\
    & The generic name is \ul{glimepiride}. \\
    PT: &  Yup, she started me on \ul{1mg} \ul{everyday}. \\
    DR: & Do you use \ul{Insulin}? \\
    PT: & The \ul{shot}? Only my brother has to. \\
    & \\
    \end{tabular}

    \begin{tabular}{r|l} \hline
    Content Span & Medications Label \\ \hline \hline
    {\it diabetes medication} & Drug \\
    {\it Sulfonylurea} & Drug \\
    {\it Amaryl} & Drug \\
    {\it glimepiride} & Drug \\
    {\it pill} & Property:Mode \\
    {\it 1mg} & Property:Dose \\
    {\it everyday} & Property:Frequency \\
    {\it insulin} & Drug \\
    {\it the shot} & Property:Mode \\ \hline
    \end{tabular}
    \caption{An example medications task with its labels.}
    \label{table:RxEg}
\end{table}

One challenge encountered in this task was choosing a specific label when other labels are equally valid. For example, ``{\it 90 day sample}'' could be marked as the total quantity consumed, or as the duration of the drug treatment. For increasing consistency, the labelers were asked to choose the labels using the preference in the order -- {\it Dose, Frequency, Quantity, Duration, Mode}.

Additionally, patients often referred to their medication using vague descriptors, such as ``{\it the pink pill}''. In many cases, the providers are able to infer the medication from the context and so the annotators are instructed to label such mentions. 

\subsection{Conditions Task}
\label{ref:Cx}
The Conditions Task, like the other two tasks, was designed based on how the discussed condition shows up in the clinical note. Even though conditions have some overlap with symptoms, they refer to broader categories and are discussed typically using clinical terminology. Ambiguities on whether a mention is a symptom or a condition was resolved by relying on the ICD-10 Code database. The condition entities were categorized into -- Condition:Patient, Condition:Family History, and Condition:Other. The attributes of conditions mirror that of symptoms, listed in Table~\ref{table:SxAttributes}, plus an additional tag to capture the onset of the condition, {\it Prop:Onset/Diagnosis}. An example of this task is illustrated in Table~\ref{table:CxEg}.

\begin{table}[t]
    \centering
    
    \begin{tabular}{ll}
    DR: & Any history of \ul{diabetes}? \\
    PT: & I have \ul{diabetes}. \\
    DR: & When was that diagnosed? \\
    PT: & \ul{10 years ago}. \\
    DR: & OK, and it seems to be \ul{well-controlled}. \\
    & Any history of \ul{high blood pressure} in the family?\\
    PT: & My brother has \ul{early onset} \ul{high blood} \\ 
    & \ul{pressure}. \\
    & \\
    \end{tabular}

    \begin{tabular}{r|l} \hline
    Content Span & Conditions Label \\ \hline \hline
    {\it diabetes} & Condition:Patient \\
    {\it 10 years ago} & Property:Time of Diagnosis \\
    {\it well-controlled} & Property:Severity \\
    {\it high blood pressure} & Condition:Family History \\
    {\it early onset} & Property:Time of diagnosis \\ \hline
    \end{tabular}
    
    \caption{An example conditions task with its labels.}
    \label{table:CxEg}
\end{table}

One challenge in the Conditions Task is that there were only a few mentions of conditions in each conversation, causing labelers to overlook the mentions inadvertently. This was mitigated with an automated method to improve recall as discussed in Section~\ref{sec:label_suggestions}.

\subsection{Relations}
\label{sec:relations}
Once all entities and attributes have been tagged and extracted from a conversation, the attributes need to be linked to their associated entities in order for them to be useful in downstream applications such as filling out sections of a clinical note.
\begin{figure}[ht]
    \centering
    \includegraphics[width=0.3\textwidth]{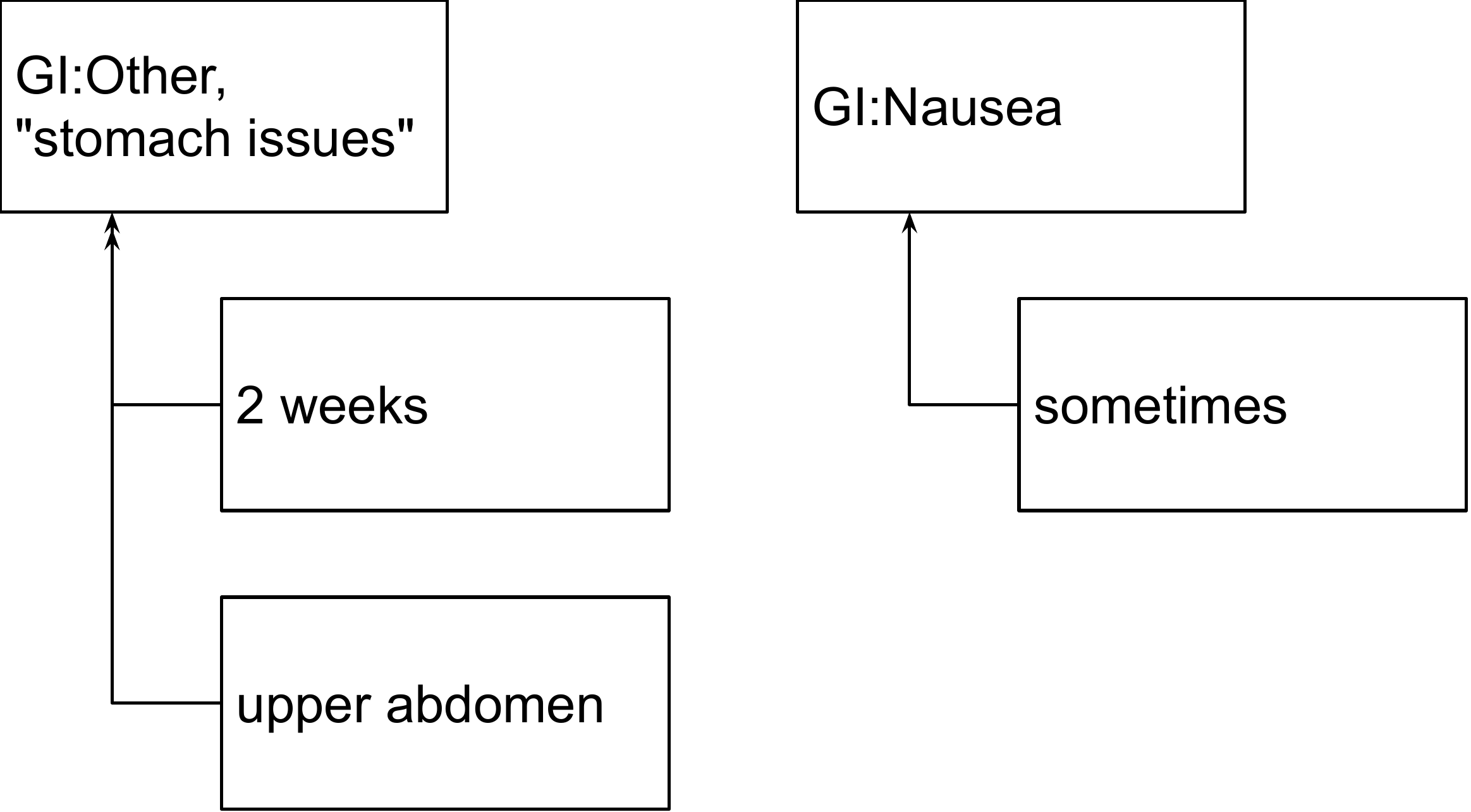}
    \caption{Illustration of relations for the example on the Symptoms Task.}
\end{figure}

In this example, the duration and location are associated with ``{\it stomach issues}'' while the frequency attribute is associated with ``{\it nausea}''. This would allow us to complete the following information in the note:
\begin{table}[h]
    \centering
    \begin{tabular}{l}
    Patient experienced \ul{stomach issues} for \ul{2 weeks} \\ in the
    \ul{upper abdomen}. Patient also experienced \\ \ul{nausea} \ul{sometimes}.
    \end{tabular}
\end{table}

One difficulty we encountered in specifying links between entities and their attributes arose because of synonymous mentions. In the example illustrated in Table~\ref{table:RxEg}, the ``{\it 1 mg}'' ``{\it pill}'' is equally related to ``{\it Sulfonylurea}'', ``{\it Amaryl}'' and ``{\it glimepiride}''. The annotators sometimes constructed an edge between the attributes and each of the drugs, and at other times chose one drug to be a canonical reference and create co-references for the other mentions of the drug. This makes it difficult to  estimate the model performance reliably. The models developed for extracting relations using these annotations are described elsewhere~\cite{Du2019b}.

\section{Task Iteration}
\label{sec:iteration}

\subsection{Training \& Quality Assurance}
\label{sec:qa}

Before starting a large scale annotation task, a number of steps were performed to refine the guidelines. 

The process consisted of the following steps:
\begin{enumerate}
  \item A small set of 3-5 conversations were labeled by a team of experienced labelers using the guidelines.
  \item Differences in their labels were resolved by them in an adjudication session. In case the differences were large, steps (1) and (2) were repeated. This resulted in finalizing the guidelines and creating a  reference set for instructional purposes.
  \item The guidelines were then introduced to the larger team of labelers who were instructed to perform the same labeling task on the reference set.
  \item Labelers were then scored based on agreement with the reference set. Those who scored well below the average quality score were given further training before proceeding to the task.
\end{enumerate}

Additionally, in order to maintain consistently high quality labels while labeling the training set, we developed a quality assurance process. The process consisted of the following steps:
\begin{enumerate}
    \item A team of experienced labelers created a reference set of labels for 3-5 conversations of varying complexity.
    \item The rest of labelers were assigned the conversations from the reference set.
    \item Labelers who scored highly with respect to the reference were chosen as reviewers. 
    \item The reviewers then reviewed the output of the other labelers by correcting the labels and documenting errors.
    \item The feedback was sent to the labelers to incorporate into future labeling efforts.
\end{enumerate}

Ultimately, we found that performing this K-way quality assurance process helped improve consistency significantly. 

\subsection{Automated Improvements}
\label{sec:label_suggestions}
We explored different techniques to improve label quality. During the Conditions Task, we noticed that labelers often overlooked the mention of conditions because they occur too infrequently in the conversation. Since the conditions are often mentioned in their canonical form, we utilized the Google Knowledge Graph to identify them in the transcripts~\cite{GKG}. The conditions identified by the knowledge graph were presented to the labelers as optional annotations that they can choose to use or discard. In order to minimize biasing the labelers, these optional annotations were presented only after the labelers finished their task and were in the process of submitting their conversation. In a controlled experiment, we found a 0.10 absolute improvement in recall from the labelers. Note, to avoid any potential measurement bias, this assistance was only provided while labeling the training set and not on the development and test sets where we relied labels from multiple labelers to maintain consistency as described further in Section~\ref{sec:evaluations}. Standard syntactic parsers were also investigated to pre-fill numberical values correspodning to dose, duration, frequency and quantity but was not employed in the labeling process.

We also experimented with using previously trained models to try to highlight numerical or date/time attributes, such as {\it Dose}, {\it Duration}, {\it Frequency}, or {\it Quantity}. 

\subsection{Common Challenges}
\label{sec:challenges}
While there were unique challenges for each of the tasks, a few challenges were common across them.

Despite the refinement of the ontologies and the continual training of labelers, there were still errors in labeling because the task is non-trivial and requires considerable attention. Certain residual errors were flagged using task-specific validation rules. These rules caught the most egregious errors, such as an attribute not being grouped with an entity, or not double-tagging a symptom or condition with its status.

A shared difficulty across all tasks was striking a balance between making the annotations clinically useful (i.e. developing a path to use the annotations to construct a complete HPI) while also reducing the cognitive burden for annotators. Through multiple experiments we uncovered significant challenges that could or could not be resolved with the addition, deletion, or substitution of tags in our ontologies.

The order in which clinically relevant events are described is often important in constructing a comprehensible HPI; for example, chest pain followed by shortness of breath and lightheadedness has different implications than shortness of breath followed by lightheadedness and chest pain. However, capturing this level of detail required expanding the ontology which was found to be cognitively burdensome for labelers and was left out to be addressed in future work.

Another challenge we encountered was confusion between  temporal tags -- {\it Time of Onset} or {\it Duration}), or progression information (e.g. {\it Improving} or {\it Worsening}. The words used to describe these attributes were often used in casual conversation manner. In the early version of the ontologies, the temporal information was annotated for each task differently, such as {\it SymProp:Frequency} and {\it MedsProp:Frequency}. This created confusion among the annotators. Subsequently, the temporal tags were defined uniformly across the tasks to the extent possible.

Another limitation of the annotation scheme is that they do not currently capture co-occurrence relationships between multiple entities although they might be clinically relevant. For example, ``{\it nausea}'' that a patient may experience along with ``{\it abdominal pain}''. The annotations also do not capture cross-ontology relationships, such as when a medication could be an alleviating factor for symptom or condition. This could be facilitated by showing labels from previous tasks when labeling a new task.

Even after carefully preparing before launching any labeling task, the labelers encountered novel situations that were not considered while developing the ontologies and the guidelines. This led to further refinements of the tasks including changing or adding a new tag. As a result, the portions of the data had to re-labeled, which was expensive. Analysis of the existing labels were used to guide these decisions, such as ignoring tags that appeared too infrequently, focusing on tags where there was a high level of disagreement, and looking at the distribution of labeled text for each tag.

While written domain corpora such as {\it i2b2} and {\it n2c2} also face similar challenges, the difficulty of this task is compounded by targeting a comprehensive ontology and information being spread across multiple speaker turns with large variations in how medical concepts are described by patients in the conversation~\cite{Uzuner2011,Henry2019}.

\subsection{Distributions \& Inter-labeler Agreements}
\label{sec:interrater}
\begin{figure}[ht]
    \centering
    \includegraphics[width=0.45\textwidth]{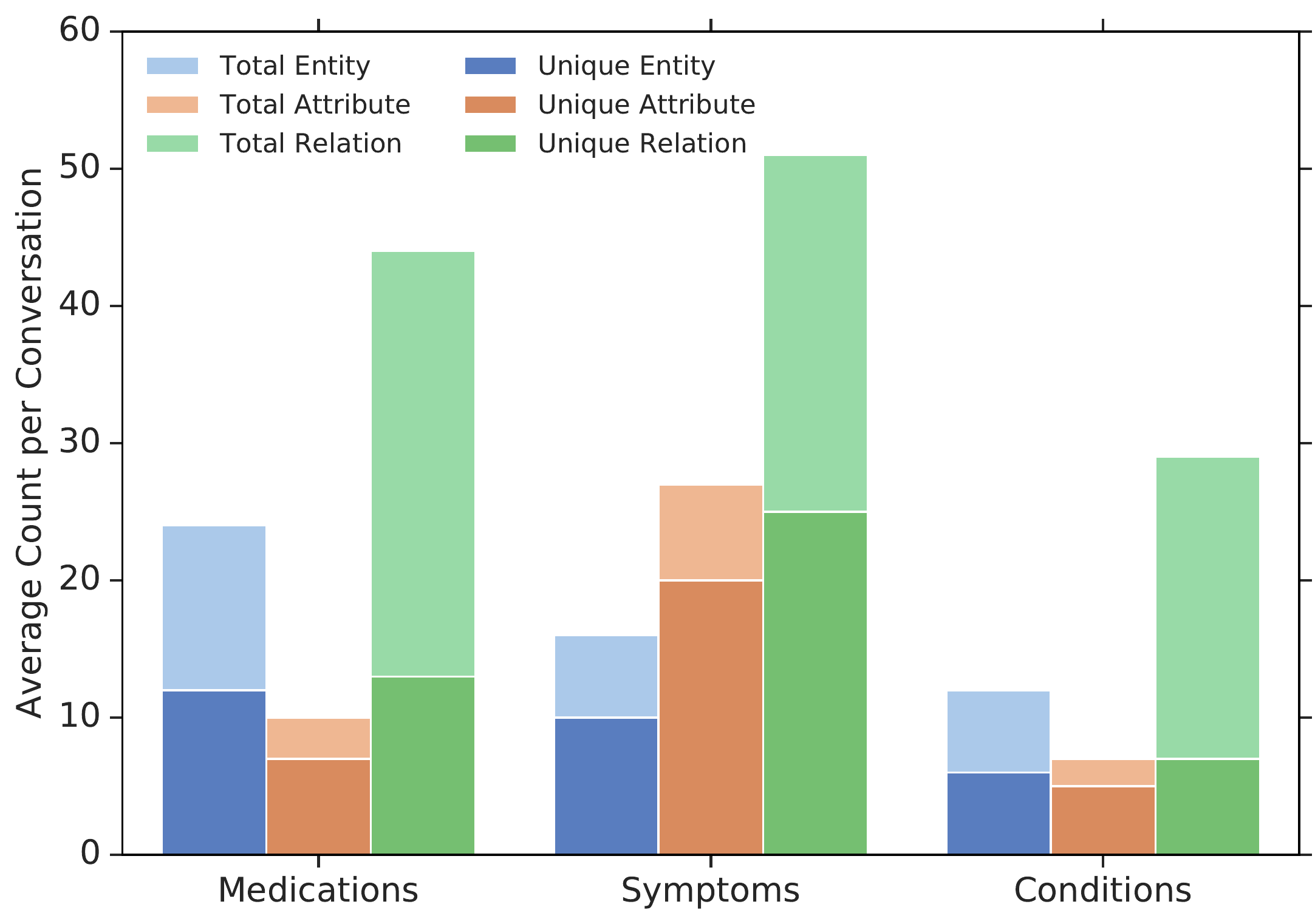}
    \caption{Numbers of labels and relations per conversation in three tasks. The unique counts are the number of unique occurrences accounting for the same highlighted span of text.}
    \label{fig:label_counts}
\end{figure}

The distribution of the occurrences of the labels in the corpus are reported in Figure~\ref{fig:label_counts}. Among entities, the number of medications mentioned is the highest. There are more attributes of symptoms than other tasks. Not surprisingly, the number of mentions of conditions and their attributes are the least among the three concepts.

\begin{figure}[ht]
    \centering
    \includegraphics[width=0.45\textwidth]{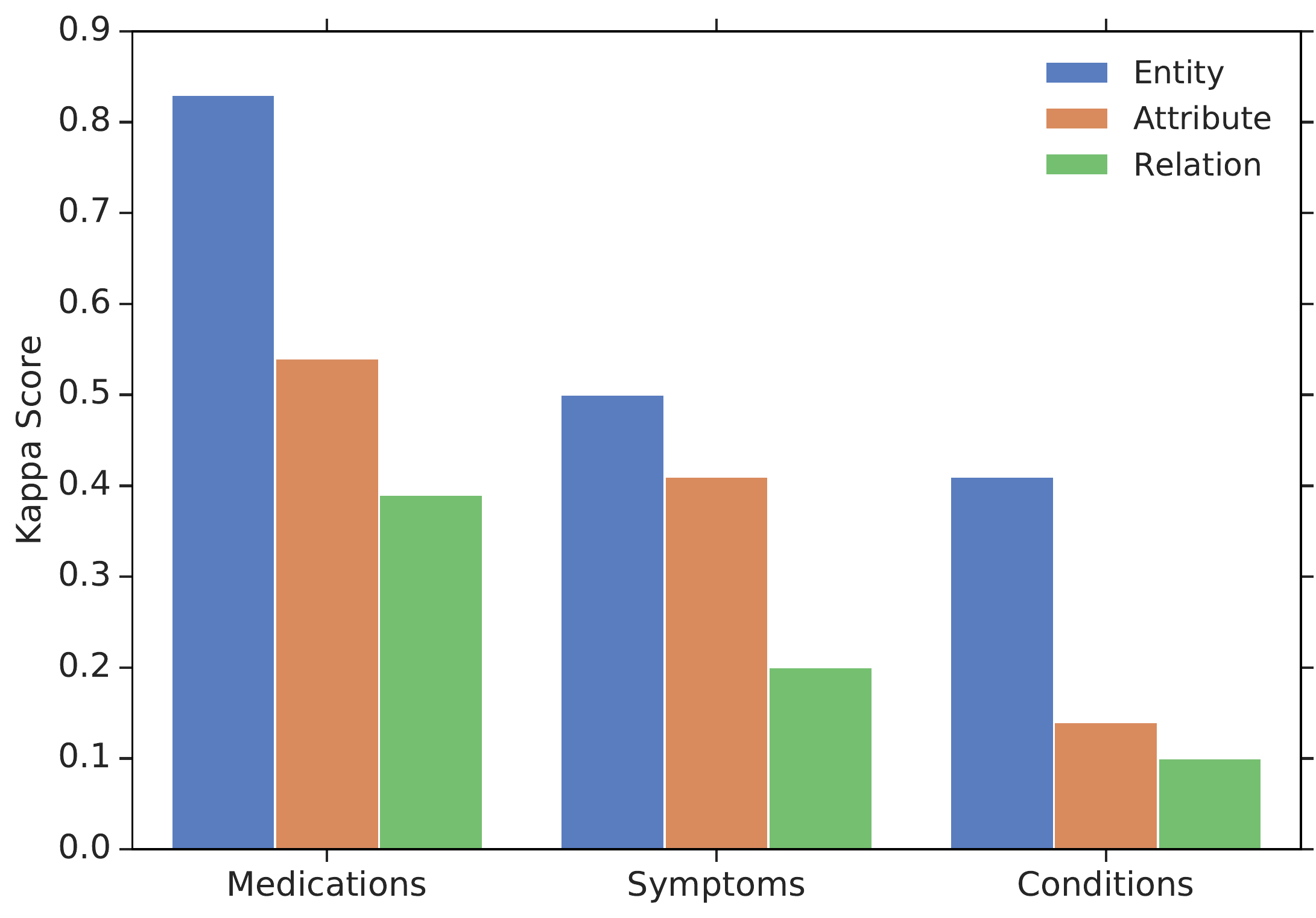}
    \caption{The inter-labeler agreement for entities, attributes, relations across three tasks.}
    \label{fig:inter_labeler_kappa}
\end{figure}

Clinical concepts vary in complexity considerably and in order to gauge the difficulty of the task and the quality of the labels, we estimated the inter-labeler agreements for the different tasks. The inter-labeler agreement was computed in terms of Cohen's kappa over the dev set of 500 conversations using three labelers per conversation. For scoring, we used a strict notion of a match that required both the label and the text content to be identical. In the Figure~\ref{fig:inter_labeler_kappa}, we can see the inter-labeler agreements are higher for entities compared to attributes or relations. Among the three tasks, the medication achieves the highest inter-labeler agreement, followed by conditions and then symptoms.

The analysis of the inter-labeler agreement helped us identify poor labels and in certain cases modifying the names of the labels to be descriptive helped improve consistency. For example, replacing {\it Response} with {\it Improving/Worsening/Unchanged} was found to be useful.

\section{The Span-Attribute Tagging Model}
\label{sec:satmodel}
One of the challenges for model development in this task is the limited amount of training data. To use the data efficiently, we developed the Span-Attribute Tagging (SAT) model which performs inference in a hierarchical manner by identifying the clinically relevant span using a BIO scheme and classifying the spans into specific labels~\cite{Du2019a}.
The span is represented using the latent representation from a bidirectional encoder, and as such has the capacity to capture the relevant context information. This latent contextual representation is used to infer the entity label and the status of the symptom as experienced or not. This is a trainable mechanism to infer status in contrast to {\it ad hoc} methods applied in negation tasks in previous work in clinical domain~\cite{Huang:2007}.

For extracting relations, we extended it to Relation-SAT model which uses a two stage process: in the first stage the candidate entities are stored in a memory buffer along with its latent contextual representation and in the second stage a classifier tests whether the candidate attributes are related to the entries in the memory buffer~\cite{Du2019b}. In both cases the models are trained end-to-end using a weighted sum of the losses corresponding to each inferred quantity.

\section{Performance Evaluations}
\label{sec:evaluations}
\subsection{Voted Reference}
For a robust evaluation of the models, we created a single ``voted'' reference from the 3-way labels mentioned in Section~\ref{sec:interrater}.  

Directly applying a standard voting algorithm is not particularly useful in improving the reference. Consider the BIO labels from three labelers, shown in Table~\ref{table:voting}, for the phrase ``{\it The pain medicine}''. While all 3 labelers agree that a part of the phrase should be labeled, they have chosen different spans. A direct voting on the BIO annotations results in assigning ``O'' to the first token, and randomly choosing the second and third tokens. 

\begin{table}[h]
    \centering
    \begin{tabular}{llll} 
    & {\it The} & {\it pain} & {\it medication} \\ 
    Labeler \#1 & Drug$_B$ & Drug$_I$ & O \\
    Labeler \#2 & O & Drug$_B$ & Drug$_I$ \\
    Labeler \#3 & O & O & Drug$_B$ \\ 
    \end{tabular}
    \caption{An example to illustrate the ``voting'' algorithm.}
    \label{table:voting}
\end{table}

We address this issue using a Markov chain, which assigns the task tags separately from the BIO notation. For each time point $t$, let $Y^1(t)$ denote the tag assigned (e.g. {\it Drug}, {\it Frequency}, etc.) and $Y^2(t)$ denote the corresponding BIO notation (e.g. B, I, O). Each token can be represented by the tuple ($Y^1(t),Y^2(t)$). We explicitly define the state at each $t$ as $Y^1(t-1)$ (with $Y^1(-1)$ defined as the empty set) and determine $Y^2(t)$ by picking the tags that maximize $P(Y^2(t)|Y^1(t), Y^1(t-1)$), where the probabilities are estimated empirically from the data. Applying this to the example above results in the annotation (,O), ({\it Drug}, B), and ({\it Drug},I), which is a more reasonable outcome than the naive voting method.

\subsection{Relaxed F-score}
\label{sec:relaxedFscore}
The annotators themselves are not always consistent in the span of text they label. E.g., ``{\it pain medication}'' vs. ``{\it the pain medication}''. To allow a certain degree of tolerance in the extracted span, the model performance was evaluated using a weighted F-score.

Let $Y_i(t)$ denote the parsed ground truth labels such that $i$ corresponds to the label index and $t$ corresponds to the token index within the sentence. We define $R_i=1/|Y_i| \sum_{Y_i} I(Y_i(t)=\hat{Y}^1(t))$ to represent the label specific recall score for label $i$, where $\hat{Y}^1(t)$ represents the model predicted tag for time point $t$. The overall recall is estimated as $R=1/n \sum_{i=1}^n R_i$ where $n$ represents the total number of ground truth labels. Similarly, label specific precision is estimated as $P_j=1/|\hat{Y}_j| \sum_{\hat{Y}_j} I(\hat{Y}_j(t)=Y^1(t))$, where $\hat{Y}_j$ represents the $j$th constructed label predicted by the model, and the overall precision is estimated as $P=1/m \sum_{j=1}^m P_j$, where $m$ represents the total number of predicted labels. The F1 score is subsequently calculated using the standard formula $2*P*R/(P+R)$. Note that updating the label specific recall and precision scores to be the cumulative product (rather than the average) results in the well known strict scores (as they then enforce that the entire span to match, rather than just a subset of the span). 

\subsection{Model Performance Analysis}
\label{sec:performance}
The performance of the SAT model on the three tasks are reported in Table~\ref{table:model_perf}. Because the Symptoms task involved a closed set of target classes for the entities, we evaluate performance by computing metrics at the conversation level ignoring the duplicates (e.g., symptoms repeated with the same status). The attributes were modeled using a common separate model. 

\begin{table}[h]
    \centering
    \begin{tabular}{|l|c|c|} \hline
    \multicolumn{3}{|c|}{Performance F1 (Precision, Recall)} \\ \hline
                & Entities & Entities+Status \\ \hline \hline
    Symptoms    & \textbf{0.72} (0.78, 0.66) & \textbf{0.60} (0.65, 0.56) \\
    Conditions  & \textbf{0.57} (0.61, 0.54) & \textbf{0.52} (0.56, 0.49) \\
    Medications & \textbf{0.90} (0.94, 0.87) & --- \\ \hline
    \end{tabular}
    \caption{The performance of the SAT model for entities in the Symptoms, the Conditions, and the Medications tasks.}
    \label{table:model_perf}
\end{table}

\subsection{Manual Error Categorization}
\label{sec:hea}
The key aim of manual error analysis is to supplement automatic evaluation of model performance by providing greater insight into the types of errors made by the model, the clinical relevance of the model errors, and plausible reasons why an error may have occurred. This analysis also allowed us to tease apart errors easily fixed in future model iterations with those that will be more difficult to correct.

Based on expected model output and anomalies in our dataset, we defined two major categories -- \emph{ErrorCause} and  \emph{ErrorImpact}. The raters were asked to utilize the context to provide their best guess on why the model might have caused the error. If they were uncertain, they could mark the cause as unknown.  Likewise, the \emph{ErrorImpact} assesses the relevance the error would have if the extracted information were populated into a clinical note. To mark an error as `\textit{relevant}', the scribe was asked to judge whether the error would actually go into the medical note. For each type of error -- Deletion, Insertion, and Substitution -- we asked the raters to attach the labels associated with \emph{ErrorCause} and \emph{ErrorImpact}. The subcategories associated with these two major categories are listed in Table~\ref{table:error_categories}. The list includes cases such as model being correct, the inferred span being incorrect, the failure of the model to use the contextual cues, errors that are associated with needing additional medical knowledge not evident from the context, and attributes that may not be related to any relevant entities.

\begin{table}[h]
    \centering
    
    \begin{tabular}{|l|l|} \hline
    Category & Subcategory \\ \hline \hline
    \multirow{9}{*}{\emph{ErrorCause}} & 
    Agree with model \\
    & Incorrect span \\
    & Ambiguous tag \\
    & Irrelevant attribute \\
    & Fail to use context \\
    & Need clinical expertise \\
    & Break in conversation flow \\
    & Clinically equivalent \\ 
    & No clear reason \\
    \hline
    \multirow{3}{*}{\emph{ClinicalRelevance}} & 
    Relevant \\
    & Not relevant \\
    & N/A \\ \hline
    \end{tabular}
    \caption{Categories and subcategories of the model errors.}
    \label{table:error_categories}
\end{table}

The errors were categorized by a group of about 10 medically trained raters, including physicians, physician assistants and professional medical scribes. Raters were trained through self review of guidelines, live instruction with guided annotation, and practice conversations. Results of the practice conversations were compared to a conversation that had been evaluated and adjudicated by a group of three individuals who had created the task guidelines. Raters were provided a side by side view of their submission compared to the adjudicated conversation. Those that showed major deviation from the adjudicated conversation were given additional guidance by task managers.

\begin{figure}[h]
    \centering
    \includegraphics[width=0.45\textwidth]{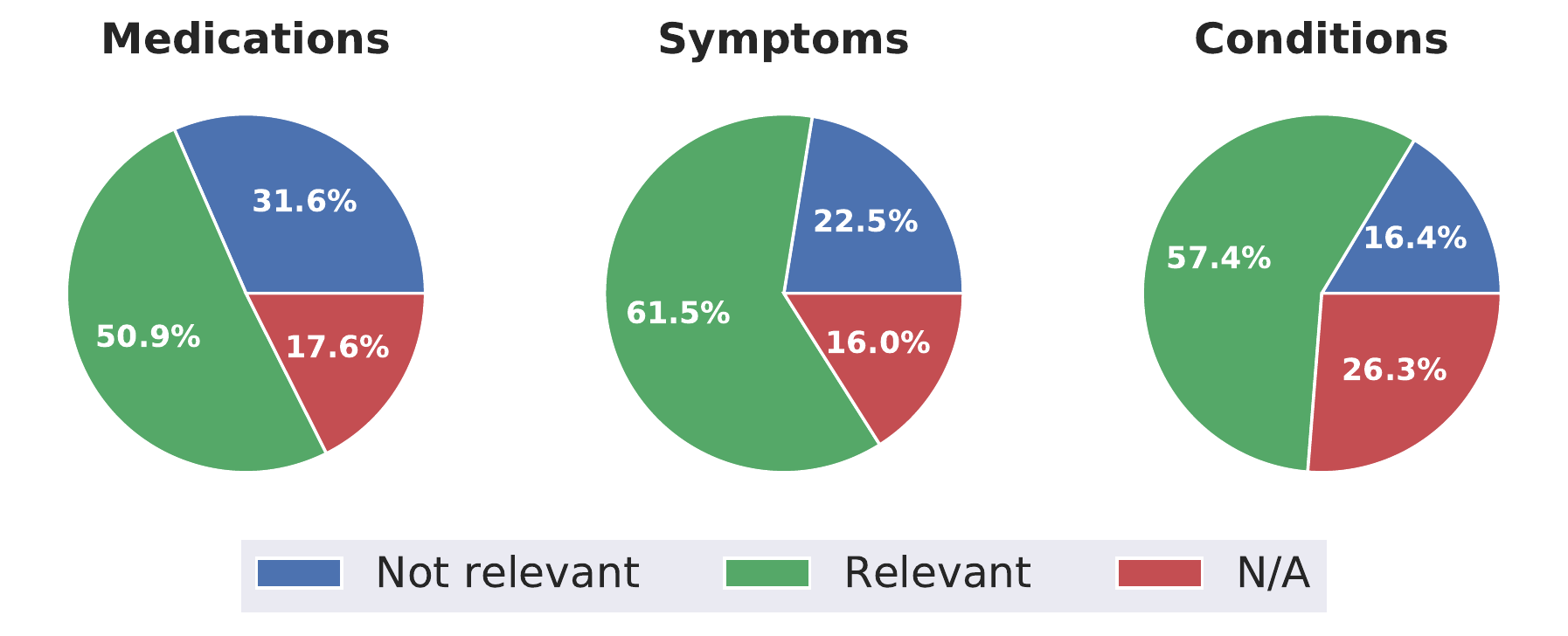}
    \caption{Proportion of errors relevant to the clinical note.}
    \label{fig:error_relevance}
\end{figure}

Among the errors committed by the model, about 17-32\% of the errors do not impact the clinical note, as shown in Figure~\ref{fig:error_relevance}. 

\begin{figure}[h]
    \centering
    \includegraphics[width=0.45\textwidth]{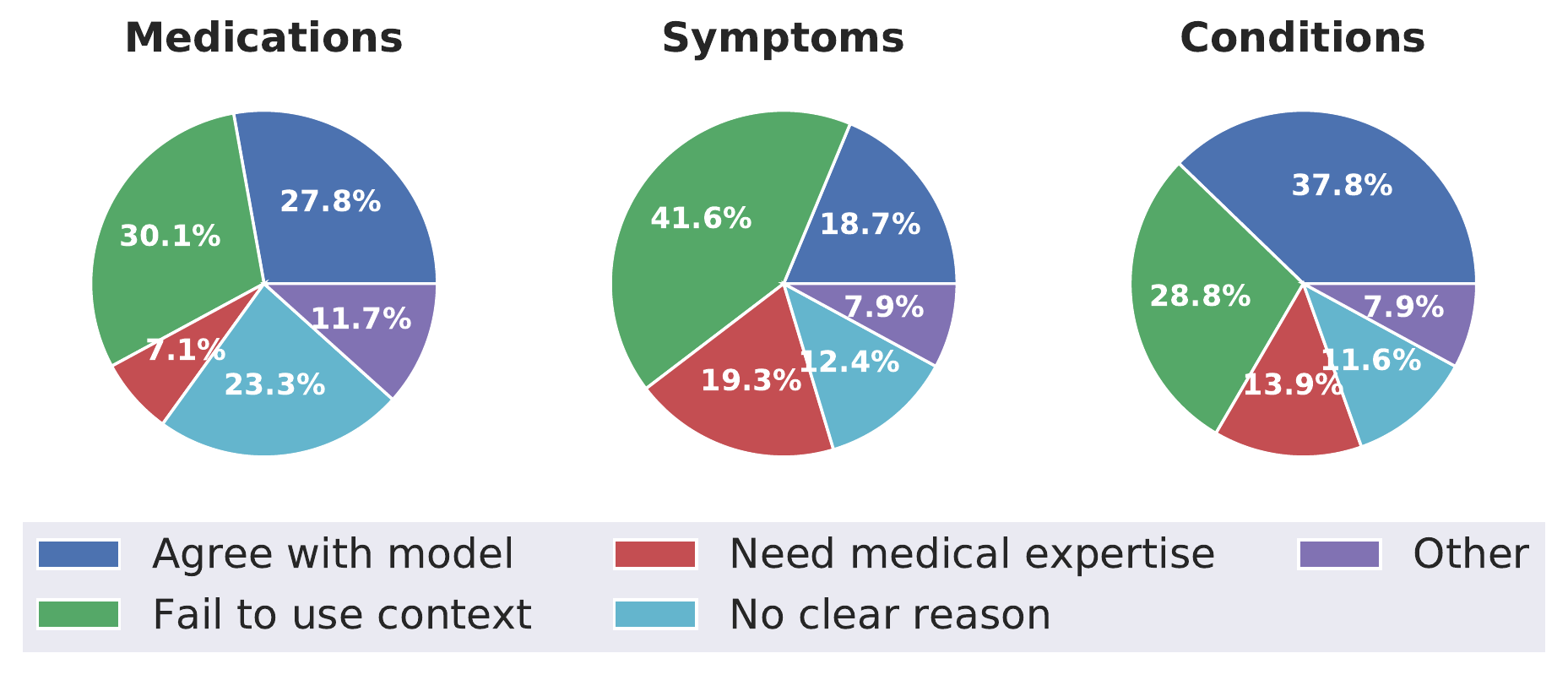}
    \caption{Proportion of errors attributed to causes.}
    \label{fig:error_categories}
\end{figure}

Human raters found that the prediction from the models were correct in about 19-38\% of the cases and that the errors were associated with the labels in the reference, as shown in Figure~\ref{fig:error_categories}. 

In analyzing the cases where the rater agreed with the model, we noticed that (despite using the voted reference labels) label quality issues still remained. Specifically, there was a noticeable amount of occurrences where the annotator missed or mis-labeled the documents. Taken together, the output of our model is significantly more useful in generating the clinical note than what is reflected in the F-score. 

Among the other errors, in about 29-42\% of the cases, the model did not take the context into account as much as humans do. For example, a patient might be describing the medication adherence as described in the example below. A human would understand that the doctor is referring to another medication that the patient is on. However, the model fails to capture this. 
\begin{table}[h]
    \centering
    \begin{tabular}{ll}
    DR: & Let's talk about your medications. How much of \\
    & the Levemir are you using now? \\
    PT: & 60 units. \\
    DR: & And \ul{the other one}?
    \end{tabular}
\end{table}

About 7-19\% of the errors relate to needing medical expertise beyond what is known from the context. In the example given below, the patient's description of their symptom (``{\it hard to breathe}'') would be labeled as {\it Respiratory:Orthopnea}. However, the model picks {\it Respiratory:Shortness of Breath} which would be a reasonable choice but not sufficiently specific given the context (``{\it when I'm lying down}'') and the medical knowledge relating to it.
\begin{table}[h]
    \centering
    \begin{tabular}{ll}
    PT: & It's really \ul{hard to breathe}. It's particularly hard \\
    & when I'm lying down in bed.
    \end{tabular}
\end{table}
Future model improvements need to focus correcting these two categories of errors.

\section{Turn Detection Model}
\label{sec:turnDetection}

Our empirical results show that the model that we developed performs poorly on attributes compared to entities. One possible explanation is that unlike entities, the span boundaries of attributes may be less distinct in a conversation. We investigated this by relaxing the model requirements, from identifying the span of words to detecting speaker turns containing the attributes, which happens to be similar to other previous work~\cite{Finley2018b,Lacson2006,Park2019}.

The turn detection model we investigated consists of an encoder, similar to the SAT model, which is followed by an attention layer and a multi-label softmax. The output predicts the probability of the occurrence of the attribute in the given input turn.

\begin{figure}[h]
    \centering
    \includegraphics[width=0.45\textwidth]{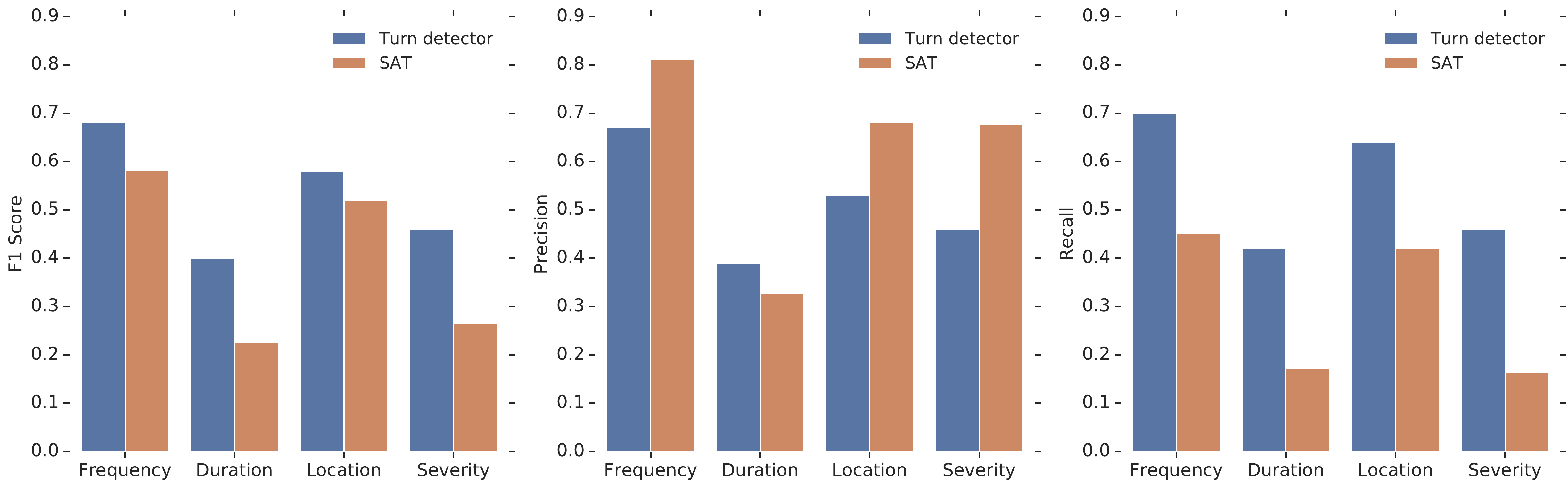}
    \caption{Comparison of performances on attributes.}
    \label{fig:model_comparisons}
\end{figure}

In the experiment, six attributes were selected as the target classes: {\it Frequency}, {\it Duration}, {\it Location}, {\it Severity}, {\it Alleviating Factor}, and {\it Provoking Factor}. These attributes were selected based on higher frequency and inter-rater agreements. Also, the model was trained and evaluated on the labels merged across all three ontologies as well as for each ontology. The model performance is reported as the F1 score, precision, recall along with our SAT model in Figure~\ref{fig:model_comparisons}. 

The result shows that the turn-based detection approach achieves better recall (but lower precision) compared to our tagging-based SAT model. The trade-off shows that when the nature of the tags does not have distinct span boundaries, modeling them at the turn-level results in better performance, especially in the situation when recall is more crucial. Note, the turn detection model was trained by treating each speaker turn as an independent input. Clearly, this can be improved further by encoding the whole conversation and predicting the class labels for each turn, which should also improve the per task attribute score.

One potential application of the turn-based model is to assist the human annotation process. For example, the model predictions can be used to pre-select a set of turns that have high probability of containing targeted tags in order to help the annotators narrow down the scope to search. Or the model predictions can be used to compare against the annotators result to help capture potential tags the annotator missed. Based on our initial experiments, a combination of the above methods can improve the efficiency and quality of the human annotations. As aforementioned, such mechanism to improve labeling efficiency should not be applied to evaluation data to avoid introducing bias in performance measurement.


\section{Discussion}
\label{sec:discussion}


To our knowledge this is the first systematic effort in extracting relevant information from provider-patient conversations with the aim of populating a clinical note. The informal nature of the conversation poses considerable challenges in extending standard methods of annotating a corpus, as described in Section~\ref{sec:corpus}. 

The complexity of extracting the relevant clinical concepts from conversational speech is considerably higher than in the written domain. For example, in a comparable work, a standard CRF model was able to achieve an F-score of 0.9 in written clinical documents~\cite{Patel2018}, whereas a simple CRF model performs poorly on our task~\cite{Du2019a}. Similarly, while developing annotation guidelines, several examples presented considerable challenge in deciding how best to annotate them, a reflection of the difficulty of the task for humans and models. 

One might argue that entities such as medication can be extracted by {\it ad hoc} approaches such as relying on dictionaries and regular expressions, as in~\cite{Finley2018b}. However, such an approach would not generalize to common place references to medications such as ``{\it the shot}'' which in certain context may refer to insulin shot and in others may not be clinically relevant. In contrast, our approach generalizes to rare occurrences of the long-tailed distribution of medications. For example, our annotators captured 8,613 unique spans of medications in the documents reviewed.

Automatic methods to assist labelers proved to be useful. The labeling throughput can be further improved using the output of previously trained models, especially for entities and attributes where the model performance is sufficiently high. For example, the medications entities, the symptoms entities, the locations and the frequencies.

Results from our manual error categorization indicate that there is room for improving the annotations further. An example that we have successfully tested is implementing a quality assurance stage where randomly sampled conversations are reviewed by senior labelers and corrected prior to being submitted for modeling. Additionally, the conversations could be down-weighted in cases where the manual annotations are substantially different from cross-validated model predictions~\cite{Wang2019}.

Another promising direction to improve the extraction of clinically relevant concepts would be to combine the SAT model and the turn detection model since they have complementary precision and recall trade-offs.

\section{Conclusions}
\label{sec:conclusion}
This paper describes a novel task for extracting clinical concepts from provider-patient conversations. We describe in detail the ontologies and the annotation guidelines for developing a corpus. Using this corpus, we trained a state-of-the-art Span-Attribute Tagging (SAT) model and report results that highlight the relative difficulties of the different tasks. Further through human error analyses of the errors, we provide insights into the weakness of the current models and opportunities to improve it. Our experiments and analyses demonstrate that several entities such as medications, symptoms, conditions and certain attributes can be extracted with sufficiently high accuracy to support practical applications and we hope our results will spur further research on this important topic.

\section{Acknowledgements}
This work would not have been possible without the support and help from a number of people, including Kyle Scholz, Nina Gonzalez, Chris Co, Mayank Mohta, Zoe Kendall, Roberto Santana, Philip Chung, Diana Jaunzeikare, and I-Ching Lee.

\newpage

\section*{References}
\bibliographystyle{lrec}
\bibliography{scribe_task}



\section*{Supplementary material}

Supplementary Materials

# The Labeling Guidelines for the Scribe Tasks

## Important Definitions

**Entity**: A medical concept that usually answers the question "what"
**Attribute**: A descriptor that usually answers "when, where, why, how, how much, how long"
**Ontology**: A collection of entity and attribute tags.
**Task**: A transcript labeling initiative focused on capturing specific information.
**Lens**: The focus around which a task is framed around.
**Guidelines (Manual)**: Instructions for how to label the conversation for a given task or tasks.
**Span (Highlight)**: The text that is captured in a highlight. Sometimes referred to as "highlight".
**Tag (Label)**: The medical terminology used to annotate a conversation.
**Group**: The coupling of at least 1 entity span and another entity span OR 1 entity span and 1 (or more) attribute spans.
**Frame**: The merging of multiple groups, that is organized by the coupling of entity spans.
**Utterances (Lines)**: A single audio sample from a speaker.
**Workspace**: Task specific labeling tool that contains conversation transcripts + audio.

## Goal

The goal of each labeling task is to **identify concepts that are clinically relevant and useful to construct a medical note.** This goal is achieved by **highlighting spans of text in the transcript, using the provided ontology(ies).** For example, in the Conditions task, the goal is to identify conditions, by highlighting and tagging all mentions, regardless of who mentions them or to whom they belong; additionally, the goal is to further define the conditions by utilizing the descriptor (attribute) tags. Our labeled data is exported to the modeling team to train the model to identify concepts defined by our ontology, therefore labelers must apply

tags as intended to avoid confusing the model. **The combination of tags and associated spans of text must be logical and must be true to the story (medical encounter).**

# Ontology

For each task there may be 1 or more ontologies. For a given labeling task, we are only interested in capturing the information in the provided ontology(ies).

**The ontology is a collection of labels (or tags)** which are identifiers for highlighted spans of text. When a span is highlighted, **the tag that defines that span is applied**. The highlighted span is how the tag or medical terminology may "appear" in a conversation. The tag Sym:GI:Abdominal pain might get assigned to "tummy hurts", because "tummy hurts" is translated from layman's speak to clinical terminology as abdominal pain. "Tummy hurts" does not get labeled with Sym:GI:Diarrhea, because those are two different symptoms.

The ontologies are **made up of 2 types of tags - entities and attributes**. Every time a particular entity of interest is mentioned in a transcript, that entity must be captured. For example, if the entity is apple, and you are reading a transcript about the fall harvest; every time a variety of apple is mentioned it must be highlighted and given the appropriate tag. However, if there were mention of a peach, this should not be highlighted with its appropriate tag, and certainly not given the "apple" tag.

# Applying the Right Label

When can a labeler reasonably apply a specific label?
- A specific label can ONLY be applied if the definitions of the highlighted span and tag are 1:1. If the highlighted span could mean 1 or more things, 1 or more tags should be applied (see [below](below)).

When should a labeler choose a more general label?
- A labeler should choose a more general label, like Condition:Other if the other label choices (Condition:Patient, Condition:Family Hx (Hereditary)) do not apply.

When should a labeler not highlight?
- A labeler **should not** highlight concepts not targeted by the task ontology. Concepts targeted by an ontology are always the entities (*every single mention*) and any **relevant** attributes. To be considered relevant to the labeling task, attributes must be discussed as they relate to modifying an entity; every mention does not warrant a highlight.
- If the entity concept is being used as a modifier. For example, "a-fib" by itself is a condition, but "a-fib medication" is a medication. Another common example, "I am stressed" implies the patient is experiencing the psychological symptom of stress, but "I had a stress test" is **not** in reference to the actual psychological symptom.

# Processes
## Labeling/Grouping Workflow

- General Labeling Approach
    - Scan the transcript until you find medical concepts.
    - Ask yourself: Is this concept applicable to the task at hand?
        - If your answer is yes, then highlight said information and assign the appropriate label.

- Detailed Labeling Approach
    - **Step 1: Identify and Highlight the Entity**
    - **Step 2: Apply the Appropriate Status Label**
    - **Step 3: Identify and Highlight Attributes**
    - **Step 4: Reread the Conversation and Group Synonymous Entities With Related Attributes**

- Detailed Grouping and Framing Rules
    - **Frames and Groups are anchored around one primary entity label. They should include all attributes related to that entity.**
        - Only entities of the **same entity label and status label** can be grouped together.
    - **At the entity level, keep all synonyms used to reference the same entity in an individual group.**
        - For example, in Diagnostics Task, if the doctor and patient discuss "blood sugar readings", "fingerstick", and "blood glucose" (all done at home), these are all considered synonyms (they're talking about the exact same test, just in different words).
            - Your group should include all entity mentions (and each synonym should have the same entity label for all 3 of these mentions of blood sugar monitoring).
        - If the same entity is discussed across the entire conversation, you should either create one large group to capture all entity synonyms with the same status/related attributes, or overlap each smaller group with at least one prior entity synonym mention.
            - This will ensure all related entities/attributes end up together in a frame. See these slides for graphical representation of this concept.
        - **IMPORTANT**: Sometimes the same text (e.g., blood glucose) should not be considered a synonym. This happens when it's being referred to in different contexts.
            - For example, in Diagnostics Task a discussion of blood glucose taken in the clinic would not be grouped with discussion of blood

- glucose that was taken at home. Likewise, blood glucose already taken should not be grouped with blood glucose pending.
    - **The same Attribute can be grouped multiple times to different Entities.**
        - It is possible for an attribute to describe more than one entity.
            - For example, if a patient says that they "have an upcoming mammogram and pap smear tomorrow," then "tomorrow" belongs in both mammogram and pap smear groups.
        - **Reminder**: Entities containing 2 **different** statuses are **NOT** permitted in the same frame; frames must contain ONLY entities with matching entity and status tags.
    - For highlighting and grouping tips & tricks, [see here](see here).
- <u>Note</u>: Base your label off of the text, not the audio. Some words may be written incorrectly in the text transcript. When this happens, base your label off of the written text, not what was said in the audio. In other words, if you hear "blood test", but "mud set" is transcribed, **do not** label this as you've heard it (you can document such discrepancies [here](here)).

---

# The Medications Task

## Medication Entities

The focus of this task is to capture every mention of medications and group them with any additional related attributes. You **should not** capture every mention of attributes, only those that are mentioned in relation to the medication. This model utilizes the **entire encounter** for context, so no need to consider "local context" when labeling and grouping.

Every label in this ontology should **capture a single concept.** Spans of text should be uniquely labeled (no overlapping highlights). Not all concepts have labels. The labels provided indicate the ones we are interested in.

A specific **priority rank** is given for each label. In the tool, the **labels are organized by priority** from highest to lowest (top-down). **In conflicts, choose the higher ranked label**.

### Medication Entity Labels + Applications

|      |      |                                                                                                                                                                     |
| ---- | ---- | ------------------------------------------------------------------------------------------------------------------------------------------------------------------- |
| Meds | Drug | All drug concepts. This includes prescribed drugs, vaccines, over the counter (OTC) drugs, and therapeutic hormones that are approved by a regulatory body (e.g., FDA, EMA, etc.). |

| | | |
|---|---|---|
| | | Examples: "Glucophage", "metformin", "flu shot", "diabetes medication", "big white oval pill", "medication", "pill for diabetes", "blood pressure medicine"<br><br>**Do not** highlight pronouns such as "it", "this", "that", "these", "those"<br><br>Priority:100 |
| Meds | Non-Drug/Supplement | All non-drug concepts and supplements. These are chemical or biological substances being **used with the intent for treatment**, but are not typically considered drugs. If these substances are mentioned outside of the context of using it as a treatment, **do not** highlight.<br><br>Examples: "Supplemental oxygen," "Activia", "Probiotic", "yogurt", "Emergen-C", "parfait", "cranberry juice", "B12", "fish oil", "folate", "vitamins"<br><br>Priority:95 |
| Meds | All Meds | General reference to **all medications taken by the patient or discussed**. Used as an entity when the doctor asks about refills in general.<br><br>Examples: "all my meds", "everything", "all of them"<br><br>When it is not clear that the mention is referring to all the medications that the patient is taking or it is referring to some medications that the patient is taking, use Meds:Drug.<br><br>PT: "Can I get refills for everything?"<br>DR: "Sure."<br>● "get refills" = MedsEvent:Continue<br>● "everything" = Meds:All Meds<br><br>Priority:80 |

# Medication Attributes

We are not focused on capturing and labeling every attribute mentioned. These labels are to be used solely for relevant mentions of medication attributes **as it pertains to the patient's experience or physician's intent.** Therefore we only want to capture attributes that accurately reflect the true dosing, frequency, status, events, etc. of an entity.

## Medication Status Labels + Applications

Affirmation of whether or not the patient is taking or not taking the medication. Answers: Is the medication actually being used?

| MedsStatus | Taking, Current/Past | Use this label if the patient affirms taking a medication. |
|---|---|---|
| | | Highlight the evidence that a patient is taking medication (If they respond "Yeah." to "Are you taking…", just highlight the response, "yeah", instead of "taking"). |
| | | If there's not a question, just a statement that the patient is taking something, you can highlight the verb "taking." |
| | | **Do not** use this for hypothetical situations. For example: <br> PT: "Only if I take ibuprofen I might get an upset stomach." <br> • "Take" = no label |
| | | **Do not** use to capture administration of vaccines/immunizations (only MedEvent tags should be applied to these entities). |
| | | **Do not** use for clarifications on attributes of the medication unless it reaffirms the true status of the medication, as well. |
| | | Example: <br> DR: "Are you taking metformin? <br> PT: "Yes." <br> DR: "Are you taking the 500 mg?" <br> PT: "No, I'm not taking 500 mg anymore, I am taking 1000 mg." <br> • "metformin" = Meds:Drug <br> • "Yes" = MedsStatus: Taking, Current <br> • "No" = no label <br> • "not taking" = no label <br> • "taking" = MedsStatus: Taking, Current <br> • "1000 mg" = MedsProp:Dose |
| | | Example: <br> DR: "So Med B… still taking it regularly I see." <br> PT: "Yeah that's right." <br> • "Med B" = Meds:Drug |

- "regularly" = MedsProp:Frequency
- "Yeah" = MedsStatus:Taking, Current

Example:
PT: "At our last visit I was still taking Med A."
- "Med A" = Meds:Drug
- "was still taking" = MedsStatus:Taking, Past
    - This sentence references the past ("In our last visit") and thus we should determine the status ("was still taking") of Med A relative to the past encounter rather than the current encounter.

Example:
DR: "Last week, you were taking the Bactrim?"
PT: "Yes."
- "Bactrim" = Meds:Drug
- "Yes" = MedsStatus: Taking, Past

As long as the patient is/was taking the medication, you can use this label for the following situations:
- Patient is fully adherent.
- Patient is partially adherent*.
- Patient is taking PRN (as needed) medication.
- Patient is taking OTC medication that is not explicitly prescribed.

***Partial adherence** includes medications not taken up to **two weeks**. If a patient stops taking their medication (forgets to pick up Rx, has meds stolen, etc) but still intends to take medication this should be labeled as MedsStatus:Taking, Current. If the lapse in medication is two weeks or greater this is considered MedsStatus:Not Taking, Current. See Example:

DR: "So you just stopped taking your Lisinopril?"
PT: "No. I just went on a three-day cruise and forgot the bottle. I started taking it again yesterday."
- "Lisinopril" = Meds:Drug
- "taking" = MedsStatus: Taking, Current

Priority: 65

| MedsStatus | Not Taking, Current/Past | Use this label if the patient affirms that they are **not taking** a medication. <br><br> Highlight the evidence that the patient is not taking the medication (so if they say "Yeah." to "You're not taking it, right?", just highlight the response, "Yeah.", instead of "not taking"). <br><br> If there's not a question, just a statement that the patient is not taking something, you can highlight "not taking". <br><br> Example: <br> DR: "Just to check, are you taking ==Med B==?" <br> PT: "==No==, I'm not." <br> - "Med B" = Meds:Drug <br> - "No" = MedsStatus:Not Taking, Current <br><br> This label should be used for every instance that the patient not taking a medication is mentioned, including: <br> - Patient was never prescribed the medication and thus is not taking it. <br> - Patient stopped medication on their own volition (a.k.a. not adherent). This can be true if the patient never started or stopped early and thus is not taking relative to the time point.** <br> - Patient finished the course of medication and stopped medication. <br><br> **Not taking is reserved for any **lapse in medication that is 2 weeks or greater**. If the patient forgets to take their medication for a few days this is still considered "taking." See Example: <br><br> DR: "You ==aren't taking== the ==Wellbutrin== then?" <br> PT: "No, it's just that I'm in a shelter right now and all my pills got stolen last month and I haven't had a chance to come in to get a new prescription." <br> - "aren't taking" = MedsStatus:Not Taking, Current <br> - "Wellbutrin" = Meds:Drug <br><br> Priority: 65 |
|---|---|---|

## Medication Property Labels + Applications

Applied to inherent properties of the medication. Answers questions: *When? Where? How? How much?*

| MedsProp | Dose | Dosage is the **amount or size** of the drug taken at a single time point or the amount or size of the individual mode (see below example 1). Include both numeric value + units if provided (e.g. 30 mg, 10 units). Dosage can also be a descriptive quantity (e.g., "very big", "almost nothing", "low"). <br><br> Examples: "500mg", "small amount", "just a little", "dose is like nothing", "0.1%" (ophthalmic solution), "Extra Strength" <br> DR: "How many ==mg== of ==metformin== are you on?" <br> PT: "I'm ==taking 500==." <br> • Mg = MedsProp:Dose <br> • Metformin = Meds:Drug <br> • "Taking" = MedsStatus:Taking, Current <br> • "500" = MedsProp:Dose <br><br> Takeaway: You may occasionally encounter dosages in which the units and dosage amounts appear on separate lines. When this occurs, label **both** the unit type and numeric value separately as MedsProp:Dose. <br><br> Of note: For this task, we **only** want to highlight and label the medication doses that the patient is **actually taking**. <br><br> Priority: 100 |
|---|---|---|
| MedsProp | Frequency | Frequency is the **interval timing** that defines the rate of medication delivery per unit time (i.e., "When?"). It conceptually maps to QD, BID, TID, QID, Q#H, QHS, PRN, etc. <br><br> Examples: "Once a day", "every four hours", "every other day", "every morning", "nighttime when you sleep", "breakfast, lunch, and dinner", "every Sunday", "a day" |

| | | |
|---|---|---|
| | | Of note: For this task, we **only** want to highlight and label the frequency that describes how the patient is **actually taking** the medication.<br><br>Priority: 100 |
| MedsProp: | Quantity | Quantity is the **count of units** of a predefined dosage (i.e., "how many?") for an individual medication. Include **both** numeric value + units (e.g. "half a tablet"). Quantity can also be a **descriptive** quantity (e.g. "I have [tons] left").<br><br>Examples: "10 pills", "ten", "just a little" "one bottle", "four pumps", "1 syringe"<br><br>DR: "I'm going to write you for a 3-month supply and you just take 2 pills once a day."<br>● "3-month supply" = MedsProp:Quantity<br>● "2 pills" = MedsProp:Quantity<br>● "Once a day" = MedsProp:Frequency<br><br>Takeaway: "supply" indicates a total amount/quantity of medication. If the DR said "I am going to write you for 3 months." then we would label "3 months" as duration.<br><br>Of note: If a patient notes he/she has "run out" of a medication, this **should not** be labeled as a quantity, as it indicates an action instead of a noun.<br>Quantity **should not** be used to capture the number of different medications (e.g. I'm on "2" blood pressure medicines).<br><br>Priority: 90 |
| MedsProp: | Duration | Duration is the **total length of time** across which all doses of medication are delivered (i.e., "How long?"). This could correlate to the total time that the patient has been taking the medication, the length of time that the doctor wants the patient to take a medication, or the length of time indicated on the prescription.<br><br>Examples: "Two weeks" total, for "three months", "years and years", "4 cycles" of |

|  |  | chemotherapy, "about a month", "a month's worth"<br><br>DR: "So let's see…you were started on the antibiotics uh how long ago?"<br>PT: "It's been 10 days, but my symptoms haven't resolved.<br>DR: OK, so I am going to write you for another 5 days and that should clear it up."<br>- "Antibiotics" = Meds:Drug<br>- "10 days" = MedsProp:Duration<br>- "5 days" = MedsProp:Duration<br><br>Priority: 85 |
|---|---|---|
| MedsProp: | Mode | Mode (A.K.A. Route of Administration) conceptually maps to PO, IV, SQ, PR, topical, buccal/sublingual, vaginal, ocular, etc. (i.e., "how?").<br><br>Examples: "by mouth", "subcutaneous", "pill", "patch", "through the needle"<br><br>Reminder: When mode is paired with quantity (e.g. "10 pills"), the entire span should be captured as quantity and **not** split up into quantity + mode.<br><br>DR: "And I'm going to give you the shot today."<br>PT: "Great."<br>DR: "Your Invega Sustenna shot."<br>- "give" = MedsEvent:Continue<br>- "the shot" = Meds:Drug<br>  - (for this "shot," the highlight answers "What?" and is therefore better classified as an entity).<br>  - (when mode is used in place of the medication entity label the text as shown above, include the word "the" if stated).<br>- "Invega Sustenna" = Meds:Drug<br>- "shot" = MedsProp:Mode<br>  - (for this "shot", the highlight answers "How?" and is therefore best classified as an attribute). |

| | | |
|---|---|---|
| | | Priority: 80 |
| MedsProp: | Instructions for Use | **Only** use for instructions not captured by Dose, Quantity, Frequency, Duration, or Mode, which would appear on the back of the box of a medication you buy at the pharmacy.<br><br>This includes **instructions to take medicine:** "Take it with food", "with your breakfast food", "do not take this medication within 2 hours of eating dairy products" and **location on body:** "under your breasts", "spread it on a hairless part of your skin."<br><br>**Do not** use this label for general instructions, education about the medication, medication mechanism of action, or other information that would not appear on the prescription label. **Do not** assume that instructing a patient to take a medication at meal time is synonymous with taking a medication with food. See Example 8.<br><br>PT: "You told me to take it on a full stomach."<ul><li>"on a full stomach" = MedsProp:Instructions for Use</li></ul>DR: "Don't take this for more than 5 days in a row."<ul><li>"Don't take this for more than 5 days in a row." = MedsProp:Instructions for Use</li></ul>Priority: 75 |
| MedsHighlight | Ambiguous | Should be applied as a **double tag** to any highlighted span that could be given multiple labels. This should also be considered when you need to refer to the priority rankings in determining the appropriate tag.<br><br>Example:<br>DR: "I'm going to give you the shot today."<br>PT: "Oh my Invega shot, ok yeah."<ul><li>"the shot"=Meds:Drug; MedsHighlight:Ambiguous</li></ul> |

| | | In the above example the first mention of shot could technically be mode, but given the context and priority rankings we label it as Meds:Drug. |
| --- | --- | --- |
| | | Example:<br>DR: "Ok so you take the Lisinopril once a day. I'll give you enough for 90 days."<br>PT: "Oh good, less visits to the pharmacy."<br>• "90 days"=MedsProp:Quantity; MedsHighlight:Ambiguous<br><br>In the above example, 90 days could be considered MedsProp:Duration, but is more accurately defined as MedsProp:Quantity, which is also supported by higher priority ranking in the Medications Task V2 guidelines. |

## Medication Relationship Labels + Applications

Applied to other medical entity concepts that have a relationship to the medication. Answers the questions: *Why? What can possibly happen? What caused the consequence?*

| MedsRel: | Benefit Experienced | Patient **confirming** benefit experienced from taking the medication.<br><br>Highlighted text is often an entity from a different label class {Sym, Condition, Lab Value}.<br><br>Highlight the actual entity that is being benefitted (e.g., headache, pain, high blood sugar), **not** the phrase that indicates that it's being benefitted (e.g., don't highlight "helps", "controls", etc.).<br><br>DR: "So the ibuprofen helps your back pain?"<br>PT: "Yes."<br>• "back pain" = MedsRel:Benefit Experienced<br><br>Priority: 35 |
| --- | --- | --- |
| MedsRel: | Benefit Not Experienced | Patient confirms that a benefit is **not** experienced from taking the medication. |

|  |  | Highlighted text is often an entity from a different label class {Sym, Condition, Lab Value}.<br><br>Highlight the actual entity that is not being benefitted (e.g., headache, pain, high blood sugar), **not** the phrase that indicates that it's not being benefitted (e.g., don't highlight "doesn't help", "doesn't control", etc.).<br><br>PT: "Metformin doesn't control my sugars."<br>● "sugars" = MedsRel:Benefit Not Experienced<br><br>Priority: 35 |
|---|---|---|
| MedsRel: | Side Effect Experienced | Patient confirms that a specific **negative** side effect is experienced from taking the medication. Theoretical side effects are **not** included. Highlighted span is often an entity from a different label class {Sym, Condition, Lab Value} and should not include the phrase that indicates the side effect.<br><br>Example:<br>DR: "Tell me how you've been doing since you're off the Invokana?"<br>PT: "I've had one low just being on the Amaryl."<br>● "Invokana" = Meds:Drug<br>● "low" = MedsRel: Side Effect Experienced<br>● "Amaryl" = Meds:Drug<br>  ○ Group "low" to Amaryl<br><br>Example:<br>PT: "The birth control pill made me nauseated, but I never had a problem with the patch."<br>● "birth control pill" = Meds:Drug<br>● "nauseated" = MedsRel:Side Effect Experienced<br>● "the patch" = Meds:Drug<br>  ○ We **do not** highlight "never had a problem" as MedsRel: Side Effect Not Experienced because it is not a specific entity (symptom, condition, lab value, etc.). |

| | | |
|---|---|---|
| | | Example:<br>PT: "Aricept caused a lot of my balance issues. I always felt off-balanced, so I stopped taking it. It's like something just doesn't feel right."<br>● "Aricept" = Meds:Drug<br>● "balance issues" = MedsRel:Side Effect Experienced<br>● "off-balanced" = MedsRel:Side Effect Experienced<br>   ○ We **do not** highlight "something just doesn't feel right" as MedsRel: Side Effect Experienced. Although this phrase may indicate a potential side effect, it is too vague to label.<br><br>Priority: 35 |
| MedsRel: | Side Effect Not Experienced | Patient confirms that the negative side effect is **not** experienced from taking the medication.<br><br>DR: "Does the pain med make you constipated?"<br>PT: "No, I've never had a problem with it."<br>● "Pain med" = Meds:Drug<br>● "Constipated" = MedsRel:Side Effect Not Experienced<br>   ○ We **do not** highlight "never had a problem" as MedsRel: Side Effect Not Experienced because it is not a specific entity (symptom, condition, lab value, etc.)<br><br>Priority: 35 |

| MedsRel: | Indication | **Clinical reason** medication is prescribed, typically corresponds to a theoretical clinical benefit, such as relief of a symptom or improvement of a lab value (e.g., blood sugar). Highlighted span is often an entity from different label class {Symptom, Condition, Diagnostic, Procedure, etc.}.<br><br>**Only** use this for **direct clinical benefits** - **do not** label logistical benefits such as insurance, cost, ease of use; **do not** label mechanism of action, physiology, or other educational statements (e.g., "this med tells your pancreas to make more insulin.").<br><br>PT: "Why did you put me on ==Coumadin== again?" DR: "Because you had a ==blood clot in your lung==. Coumadin works by thinning your blood which helps try to prevent you having another one."<br>● "Coumadin" = Meds:Drug<br>● "blood clot in your lung" = MedsRel:Indication<br>   ○ We **do not** HL "thinning your blood" because in this context, the doctor is explaining how the medicine works.<br><br>Priority: 30 |
|---|---|---|

## Medication Event Labels + Applications

**Verb phrase** describing a **medication change** that was **initiated by the provider**. Answers the question: *What does the provider want to happen to the medication entity and its attributes?*
- **Grammar Reminder:** A verb is the word that indicates the action. A verb phrase includes the verb AND any direct or indirect object, **but not** the subject.

| MedsEvent: | Start, Current/Past/ Future | **Verb Phrase** describing a **provider's** decision to **start taking** a medication entity. Should be used in the event that a provider is **starting** a new medication or is **restarting** a medication that has been **stopped by a provider** in the past. This can be mentioned either by the provider or patient, but it is clear that the |
|---|---|---|

| | | event was **initiated by the provider** |
|---|---|---|
| | | This **should not** be used to capture theoretical mentions of a medication start event (e.g. if decision to start is dependent on a pending lab result, worsening of condition, etc). |
| | | PT: "I started taking a supplement that my friend told me to try, but it's not helping." <br> DR: "I want you to stop that and start taking this prescription I am going to write." <br> • "taking" = MedsStatus: Taking, Current <br> • "supplement" = Meds:Non-Drug/Supplement <br> • "stop" = MedsEvent: Stop, Current <br> • "start taking" = MedsEvent: Start, Current <br> • "this prescription" = Meds:Drug <br>   ○ Note: We **do not** highlight "started taking" as MedsEvent:Start, Past because the supplement is something the patient started taking on her volition - it was not recommended by a doctor. |
| | | Priority: 75 |
| MedsEvent: | Stop, Current/Past/Future | **Verb Phrase** describing a **provider's** decision for the patient to **stop taking** a medication entity. If the patient reports stopping their medication without a provider's direction, that is Not Taking. This can be mentioned either by the provider or patient, but it is clear that the event was **initiated by the provider.** |
| | | Example: <br> DR: "Stop taking ibuprofen when the swelling goes down." <br> PT: "Ok, I'm guessing it'll only be a day or two from now." <br> • "Stop taking" = MedsEvent:Stop, Future <br> • "ibuprofen" = Meds:Drug |
| | | Example: |

| | | |
|---|---|---|
| | | DR: "I want you to stop taking that when you finish all the pills."<br>● "stop taking" = MedsEvent:Stop, Current<br><br>Example:<br>PT: "Last week you told me to stop taking St. John's Wort so I haven't taken it since."<br>● "stop taking" = MedsEvent:Stop, Past<br>● "St. John's Wort" = Meds:Non-Drug/Supplement<br>● "haven't taken" = MedsEvent:Not Taking, Current<br><br>Priority: 75 |
| MedsEvent: | Continue, Current/Past/Future | **Verb Phrase** describing **provider's** decision for the patient to **continue taking** a medication entity with unchanged medication properties. This can be mentioned either by the provider or patient, but it is clear that the event was initiated by the provider.<br><br>If **changes to attributes**, such as dose, quantity, frequency are made, **do not** use this label (use MedsEvent:Modify instead). This label **should not** be used for verbs that describe the status of whether patient is taking, not taking, or completed medication (see MedStatus).<br><br>Example:<br>DR: "That's good, continue that."<br>● continue = MedsEvent:Continue, Current<br><br>Example:<br>DR: "Ok, so the dermatologist told you to continue using the cream on your dry skin, how's that going?"<br>● "continue using" = MedsEvent:Continue, Past<br>● "the cream" = Meds:Drug<br><br>Example:<br>DR: "Do we need to refill your blood pressure medication?" |

|  |  | PT: "Yes." <br> • "need to refill" = MedsEvent:Continue, Current <br> • "blood pressure medication" = Meds:Drug <br>   ○ We highlight "need to refill" as MedsEvent:Continue, Current, because the act of refilling a medication denotes a continuation of the prescription given by the physician. <br><br> Priority: 60 |
| --- | --- | --- |
| MedsEvent: | Modify, Current/Past/Future | **Verb Phrase** describing a **provider's** decision to **modify properties** of the medication (e.g. Dose, Quantity, Frequency, Duration, Mode, Instructions for Use). This can be mentioned either by the provider or patient, but it is clear that the event was **initiated by the provider**. <br><br> Example: <br> DR: "Yes, let's increase the dose today to 50." <br> • "increase the dose" = MedsEvent:Modify, Current <br> • "50" = MedsProp:Dose <br><br> Example: <br> DR: "So your other doctor at the nursing home increased the dose two weeks ago to 50." <br> • increase the dose = MedsEvent:Modify, Past <br> • 50 = MedsProp:Dose <br><br> **Do not capture temporary changes** in a patient's medication. For instance, if a provider tells a patient to increase their daily dosage of a medication they take on a daily basis for five days, or until symptoms subside, then this information **should not** be highlighted. Only the actual prescription information should be highlighted. This applies to medications that the patient is instructed to refrain from taking temporarily. See examples below: |

|   |   |   |
|---|---|---|
|   |   | DR: "I'm going to have you double your ==Lasix== dose for the next three days. Get you really flowing. So you'll do 120 mg in the morning for the next three days."<br>    ● "Lasix" = Meds:Drug<br>        ○ The temporary increase in the medication is **not** labeled (including the temporary dosage amount, frequency, and duration).<br><br>DR: "So I want you to stop taking the Coumadin 2 or 3 days before the procedure. Then you can start it back up again right after."<br>    ● "==Coumadin==" = Meds:Drug<br>        ○ The patient has been asked to stop taking the Coumadin for just a few days. No event is really taking place and the patient will continue to take the medication as prescribed.<br><br>Priority: 60 |
| MedsEvent: | Switch, Current/Past/Future | **Verb Phrase** describing a **provider's** decision to **switch** from taking one medication entity to another medication entity. This can be mentioned either by the provider or patient, but it is clear that the event was initiated by the provider.<br><br>This label **should not** be used for verbs that describe the status of whether patient is taking, not taking, or completed medication (see MedStatus).<br><br>**When grouping**, please group this Medication Attribute to the entity that is being stopped and the entity that is being started in 2 separate groups.<br><br>DR: "I am going to ==change== your ==diabetes medication==. We are going to ==switch== you from ==Glucophage== to ==Glucophage XR==."<br>PT: "Okay."<br>DR: "So, I want you to ==stop== ==Glucophage== and ==start== ==Glucophage XR==." |

| | | <ul><li>"change" = MedsEvent:Switch, Current</li><li>"diabetes medication" = Meds:Drug</li><li>"switch" = MedsEvent:Switch, Current</li><li>"Glucophage" = Meds:Drug</li><li>"Glucophage XR"= Meds:Drug</li><li>"stop" = MedsEvent:Stop, Current</li><li>"start" = MedsEvent:Start, Current<ul><li>We **do not** use MedsEvent:Switch for "start" and "stop" because their respective labels more specifically capture the event and have a higher priority ranking.</li></ul></li></ul>Priority: 60 |
|---|---|---|

## Example 1 - Benefit vs. Indication

93    **PT**    "I take the medicine for my back pain.  It helps a lot."

| Group 1 | the medicine | Meds:Drug |
|---|---|---|
| | take | MedStatus:Taking,Current |
| | Back pain | MedsRel:Benefit Experienced |

Takeaways:
- While both are true, we can only choose a single label for "back pain".  We prioritize labeling "experienced" concepts over any theoretical benefit or side effect that may be discussed. When in doubt, refer to the priority ranking.

## Example 2 - (Not) Taking, Past vs (Not) Taking, Current

93    **DR**    "Last time this happened did they give you any medicine to help with nausea?"

94    **PT**    "No."

| Group 1 | medicine to help with nausea | Meds:Drug |
|---|---|---|

|  | no | MedsStatus:Not Taking, Current |
|---|---|---|

Takeaways:
- While you may be tempted to use the Not Taking, Past label given the med event in question was in the past, we always prioritize capturing the patient's *current status*. If you are uncertain and debating between any two labels, always refer back to the priority rankings.
- It is important to also capture the entire phrase "medicine to help with your nausea" as one label, because it is referencing one specific type of medication. As opposed to labeling medicine (Meds:Drug) and nausea (MedsProp:Indication).

### Example 3 - Quantity vs Frequency

94    **PT**    "I've been trying not to get sick all school year , so I'm drinking Emergen-C a lot."

| Group 1 | Drinking | MedsStatus:Taking, Current |
|---|---|---|
|  | Emergen-C | Meds:Non-Drug/Supplement |
|  | A lot | MedsProp:Frequency |

Takeaways:
- While the phrase "a lot" sounds like a(n) quantity/amount, in this example it is a frequency given the context of the conversation. In this context, the patient is stating that they are drinking the Non-drug/Supplement frequently as opposed to drinking a large quantity and therefore is captured as frequency.

### Example 4 - Some Meds vs All Meds

94    **DR**    "Do you need a refill for any of your medications?"

95    **PT**    "Well…before we go there, can I ask you a question about my diabetes meds?"

| Group 1 | Any of your medications | Meds:All Meds |
|---|---|---|
| Group 2 | Diabetes meds | Meds:Drug |

22

Takeaways:
- The first mention of medications is labeled "All Meds", because the doctor is referencing all medications the patient is currently prescribed. The second mention of medications is labeled "Meds:Drug", because the patient is referencing a subset of their medication; diabetes medications are only <u>some</u> of the total (all medications).

### Example 5 - Meds Event:Start vs Continue

| 92 | PT | "I stopped taking Humalog for one week, because I ran out of refills." |
|---|---|---|
| 93 | DR | "We need to get back on that." |

| Group 1 | Humalog | Meds:Drug |
|---|---|---|
| | get back on | MedsEvent:Continue, Current |

Takeaways:
- "Stopped taking" is not labeled, because the patient did not stop the medication on their own volition **AND** they have not been off their medication for 2 or more weeks.
- We label "get back on" as MedsEvent:Continue, Current, as opposed to MedsEvent:Start, Current, to indicate the physician's intent to continue the patient's medication.
  - Furthermore, there was no MedsEvent:Stop that occurred beforehand as discussed above.

### Example 6 - Chemotherapy and Radiation

| 92 | PT | "Now that I'm no longer on chemotherapy and radiation, I feel much better." |
|---|---|---|

| Group 1 | Chemotherapy | Meds:Drug |
|---|---|---|
| | no longer taking | MedsStatus:Not Taking, Current |

Takeaways:
- In this example, the patient mentions two forms of cancer treatment - chemotherapy and radiation - but only chemotherapy is a medication and therefore captured as "Meds:Drug".
  - Radiation is a procedure and is therefore **not** captured in this task.

### Example 7 - "Refills" = MedsEvent:Continue

92 **DR** "Do you need refills for anything?"

93 **PT** "Yes, my asthma medicine."

| Group 1 | Asthma medicine | Meds:Drug |
| --- | --- | --- |
| | Need refills | MedsEvent:Continue, Current |

Takeaways:
- "Need refill" is HL and labeled only because the patient actually needs a medication refill. In the event that the provider asks the patient if they need a refill and they answer "no", we **do not** label the word "refill." Additionally, if the patient requests a refill and the provider does not indicate that they are going to fulfill that request, refill similarly **should not** be labeled.
- If "need" is mentioned by itself, it could mean continue or start. It is important to rely on the surrounding context to determine whether the med event is a continue event vs a start event.

### Example 8 - Instructions for Use vs. Frequency

92 **DR** "For the medicine, I'd like for you to take one pill in the morning with breakfast, and one pill right before bed."

| Group 1 | Medicine | Meds:Drug |
| --- | --- | --- |
| | Take | MedsEvent:Start, Current |
| | One pill | MedsProp:Quantity |
| | In the morning | MedsProp:Frequency |
| | One pill | MedsProp:Quantity |
| | Before bed | MedsProp:Frequency |

Takeaways:
- The doctor is referencing actions related to the time of day or frequency of taking the medication in this context.

- The inclusion of "with breakfast" is superfluous. The doctor is not instructing the patient to take this medication with food, therefore the label instructions for use does not apply.

## Example 9 - Mode vs Drug

92  **DR**  "Are you taking any inhaler for your asthma?"

93  **PT**  "Yes, I'm taking the albuterol inhaler and I like it."

94  **DR**  "Okay go ahead and continue taking that inhaler."

| Group 1 | Inhaler | Meds:Drug |
| --- | --- | --- |
| | Asthma | MedsRel:Indication |
| | Yes | MedsStatus: Taking, Current |
| | Taking | MedsStatus: Taking, Current |
| | Albuterol | Meds:Drug |
| | Inhaler | MedsProp:Mode |
| | Continue | MedsEvent:Continue, current |
| | That inhaler | Meds:Drug |

Takeaways:
- Pills/mode/etc. can be considered either **drug or mode** based on the context in which it is mentioned.
    - When the mention is being used as a noun on its own (e.g. you can replace that HL span with the name of the actual drug and it still makes sense), then it is being used as a synonym to the drug and would be labeled accordingly as an **entity**.
    - Alternatively, when it is being used as a descriptive modifier indicating the kind of drug being taken and/or the way the drug is being taken, and **not** as a standalone noun, then it would be labeled as **mode**.

## Example 10 - Nonspecific Med Entity

92  **DR**  "Any other medications?"

| 93 | PT | "Yes, I take my thyroid every morning." |

| Group 1 | thyroid | Meds:Drug |
|---|---|---|
| | take | MedsStatus:Taking, current |
| | every morning | MedsProp:Frequency |

Takeaways:
- When a word/phrase such as "thyroid" in the above example is clearly being used to reference a medication entity, it is appropriate to capture accordingly.

### Example 11 - Instructions For Use

| 12 | DR | "What I want you to do with the cream is, apply this first thing in the morning and then if you need more, you can reapply it throughout the day, but make sure you only use it sparingly." |

| Group 1 | cream | Meds:Drug |
|---|---|---|
| | apply this first thing in the morning and then if you need more, you can reapply it throughout the day | MedsProp:Instructions for Use |
| | Only use it sparingly | MedsProp:Quantity |

Takeaways:
- Rely on the context cues to determine whether or not the provider is providing instructions for use, i.e. *how* to use the medication. The context cue in this example is, "what I want you to do". This may not always be the case, so use your best judgement and rely on your clinical exposure.

### Example 12 - "Samples"

| 40 | DR | "Okay, so this is the Repatha. I am going to give you some samples of that, okay." |
| 41 | PT | "Okay, and then what?" |

| | | |
|---|---|---|
| 42 | DR | "Well, finish the sample and then let's see how you do." |

| Group 1 | Repatha | Meds:Drug |
|---|---|---|
| | Samples (line 40) | MedProp:Quantity |
| | Sample (line 42) | Meds:Drug |

Takeaways:
- Depending on the context, a mention of a "*sample*", could imply **quantity** or the drug **entity** itself. If there is any ambiguity in the text, then you must rely on the priority ranking and assign the Meds:Drug tag. Additionally, remember to assign the Ambiguous or Difficult tag when applicable.

### Example 13 - Mentions of "Generic" or "Prescription"

| | | |
|---|---|---|
| 11 | PT | "I am taking the generic." |
| 12 | DR | "Okay, good. I will write you a new prescription for that." |

| Group 1 | Am taking | MedsStatus:Taking, Current |
|---|---|---|
| | generic | Meds:Drug |
| | New prescription | **No HL** |

Takeaways:
- When vague terms such as "generic" or "prescription" are clearly replacing a medication entity, they may be appropriate to capture, E.g. if you can replace the span of text with the actual medication name. However, if these mentions are simply acting as an adjective/modifier OR referring to something like a written/electronic prescription, **do not** capture.

### Example 14 - Conflicting Statuses

| | | |
|---|---|---|
| 67 | DR | "So have you been taking your diabetes medication?" |
| 68 | PT | "Yes, I'm still taking it." |
| 69 | DR | "Ok, good." |
| 70 | PT | "But I actually ran out 3 weeks ago, so I've been off of it." |

| Group 1 | Diabetes medication | Meds:Drug |
| | Yes | **No HL** |
| | Still taking | **No HL** |
| | Ran out | **No HL** |
| | Been off | MedsStatus:Not Taking, Current |

Takeaways:
- The patient initially states that they have been taking their medication, but then clarifies that they have not been taking it the past 3 weeks, because they "ran out". Because we only want to capture attributes that accurately reflect the patient's experience, only "been off" should be captured as MedsStatus:Not Taking, Current. The patient states they "ran out" 3 weeks ago, which is greater than the 2 week guideline we use to determine if a span should be captured as Taking or Not Taking.

### Example 15 - Vaccinations/Immunizations

| 67 | DR | "Your flu shot is up to date, but it looks like you need your pneumonia one. Are you okay to do that today?" |
| 68 | PT | "Sure." |

| Group 1 | Flu shot | Meds:Drug |
| Group 2 | need | MedsEvent:Start, Current |
| | Pneumonia one | Meds:Drug |

Takeaways:
- MedStatus tags **should not** be applied as attributes for vaccinations/immunizations. Instead, MedEvent tags should be applied to the appropriate **verb phrase** to indicate when a vaccination was or is going to be administered.

While "pneumonia one" is a somewhat vague entity, it is clearly referring to a pneumonia vaccination based on the context and is therefore appropriate to capture.

# The Symptoms Task

## Symptom Entities

The goal of this task is to accurately capture **every mention of symptoms**, designate a status for each symptom, and group with any additional related attributes.

**Symptom** = **subjective** evidence of disease or physical disturbance observed by the patient; a departure from a patient's normal functioning or feeling. Broadly: something that **indicates** the presence of a physical disorder.
- E.g. headache is a symptom of many diseases. Visual disturbances may be a symptom of retinal arteriosclerosis. Chest pain may be a symptom of a heart condition.

Symptom entity labels ("Sym") are organized by systems (e.g., CV, Resp, GI). Each symptom entity label is a specific clinical term describing a symptom.
- You should select the symptom label that best fits the concept being described in the conversation.

<u>Of note</u>: we label based on what is being conveyed in the conversation, not based on the actual words.
- Meaning, if the patient says "pain in my neck", but it can clearly be discerned that they mean "sore throat", you should favor labeling this concept based on what it means within the context of the conversation, and not as "pain".

## Symptom Attributes

Attributes that describe the Symptoms are found under Symptom Attribute labels ("SymAttr") and there are 4 categories of attributes:
1. Statuses (SymStatus)
2. Properties (SymProp)
3. Relationships (SymRel)
4. Specializations (SymSpecial)

### Symptom Attribute Labels + Applications

| SymStatus | Experienced | Symptom entities experienced by the patient. |
|---|---|---|
| SymStatus | Not Experienced | Symptom entities not experienced by the patient. |
| SymStatus | Experience Unknown | When the patient's experience with a symptom entity is unknown. |
| SymStatus | Experience, Theoretical | Symptom entities that are hypothetical or theoretical (can be educational or potential side |

| | | effects). |
|---|---|---|
| SymStatus | Other's Experience | Symptom entities that are experienced by another individual. |
| SymProp | Time of Onset | A word or phrase indicating when a symptom began. |
| SymProp | Frequency/Tempo | A word or phrase indicating how *often* a symptom occurs. |
| SymProp | Duration, All Time | A word or phrase indicating how long the patient has been with a symptom. |
| SymProp | Duration, Episodic | Words or phrases that indicate how long *an* episode of the symptom lasts. |
| SymProp | Improving | Word(s) that imply the symptom is improving. |
| SymProp | Worsening | Word(s) that imply the symptom is worsening. |
| SymProp | Unchanged | Word(s) that imply the symptom is unchanged. |
| SymProp | Location (on body) | The anatomical location, when not otherwise stated in the symptom entity label. |
| SymProp | Severity/Amount | Word(s) that imply and are confirmed by the patient to describe the severity of a symptom.<br><br>For severity, the span of text should be easily inserted into the phrase "the (symptom) is so bad that it (insert HL text)".<br><br>For amount, this should be a numerical description of the symptom entity. |
| SymProp | Characteristic/Quality | Word(s) that describe the symptom; adjectives not captured by the other attribute labels. |
| SymRel | Provoking Factor, Yes | Meds/Non-Meds/NMT/activity that is considered to be the cause of the symptom. |
| SymRel | Provoking Factor, No | Meds/Non-Meds/NMT/activity that is denied by the patient to be the cause of the symptom. |
| SymRel | Provoking Factor, Unknown | Meds/Non-Meds/NMT/activity considered to be the potential cause of the symptom, not denied or confirmed. |
| SymRel | Provoking Factor, Theoretical | Meds/Non-Meds/NMT/activity discussed by the DR that may provoke a specified symptom (not yet tried by the PT). |
| SymRel | Alleviating Factor, Helped | Meds/Non-Meds/NMT/activity that has been tried and helped a symptom. |
| SymRel | Alleviating Factor, Doesn't | Meds/Non-Meds/NMT/activity that has been tried |

| | | |
|---|---|---|
| | Help | and didn't help a symptom. |
| SymRel | Alleviating Factor, Unknown | Meds/Non-Meds/NMT/activity tried by the patient, but efficacy not specified by the patient. |
| SymRel | Alleviating Factor, Theoretical | Meds/Non-Meds/NMT/activity on which the DR is providing education or recommending, and has not yet been tried by the patient. |
| SymSpecial | Unclear Condition | For special cases when a condition is described as a symptom (described as experienced/perceivable) and it is unclear whether this warrants a label.<br><br>This tag should only be applied as an additional tag to Symptom Entities. |
| SymSpecial | Past | Symptom entities that were/were not experienced by the patient in the past. |
| SymTemp | Potential Attribute | Labeled attribute concepts that have not yet been established as pertinent to the medical encounter due to an apparent absence of associated entity.<br><br>This tag must either be replaced with the appropriate attribute tag OR removed before submitting the conversation if no associated entity is present. |

### Example 1 - Ambiguous Symptom

- **PT** "I stopped taking those meds because of multiple stomach problems."

| Group 1 | Stomach problems | Sym: GI: Abdominal Pain; Sym: GI: Nausea; Sym: GI + SymStatus:Experienced |
|---|---|---|

Takeaways:
- If the patient previously described their "stomach problems" or elaborated within 10 lines that they are specifically having diarrhea, you can reasonably apply Sym:GI:Diarrhea. If no further details are provided prior to this line nor within the next 10 lines, up to 3 Sym:GI tags should be applied to account for the multiple symptoms possibly being referenced.

### Example 2 - SymStatus:Other's Experience

12   **PT**   "My roommate had a cough for the past few days, now I have it

too."

| Group 1 | cough | Sym:Resp:Cough + SymStatus:Other's Experience |

Takeaways:
- The SymStatus:Other's Experience label should be applied to **symptoms of sick contacts.**

### Example 3 - SymStatus:Experience, Theoretical

104   **DR**   "Let's talk about this pain you've been having."

105   **PT**   "Well, it started when I began taking this new medication."

106   **DR**   "The statin."

107   **PT**   "Yeah, that statin."

108   **DR**   "Well, that is what I cautioned you about, remember. Statins have a well known side effect of muscle pain."

| Group 1 | Pain | Sym:MSK:Pain + SymStatus:Experienced |
| --- | --- | --- |
| | New medication | SymRel:Provoking Factor, Yes |
| | Statin | SymRel:Provoking Factor, Yes |
| | statin | SymRel:Provoking Factor, Yes |
| Group 2 | Statins | SymRel:Provoking Factor, Theoretical |
| | Muscle pain | Sym:MSK:Pain + SymStatus:Experience, Theoretical |

Takeaways:
- The SymStatus:Experienced, Theoretical label should be applied to **hypothetical symptoms, potential side effects, and symptoms that are theoretical benefits**.
- While the patient is experiencing this exact pain (muscle pain secondary to taking a statin), the DR's mention of the pain in line 108 is educational and is not a mention of the patient's own experience. Additionally, the mention of statin in that line is

theoretical, because they are talking about a potential side effect, which is not necessarily what the patient is experiencing.

## Symptom Entity Synonyms

- Symptom entities **do not** need to be grouped to any other concepts, **unless** there is another (synonymous) mention of the symptom entity.
    - Symptom entities you may want to group may be expressed using the same words, but may also be expressed in other words.
- Please note, if the same words are mentioned or entity concepts have the same labels (entity label + status label) this **does not** necessarily mean the two or more are synonymous.
    - We group based on the concept that is actually being conveyed within the conversation, not the label or word(s) by themselves.
        - For example, "pain" = Sym:MSK:Pain + SymStatus:Experienced, can be used for highlighted spans of text that convey completely different concepts, E.g. shoulder pain from an injury and unprovoked leg pain (these different concepts **should not** be grouped together even though the concepts have the same labels).

### Example 4 - Same Entity Labels, Different Frames

| 92 | DR | "So, how have you been? Any pain anywhere?" |
| 93 | PT | "I gotta tell you doc, it seems like I'm falling apart." |
| 94 | DR | "How so?" |
| 95 | PT | "Well, last week I fell and landed on this shoulder, so I've got pain there. Then, this knee, which is old, that pain never got any better after the surgery, even on the medication." |

| Group 1 | last week | SymProp:Time of Onset |
| | Fell | SymRel:Provoking Factor, Yes |
| | Shoulder | SymProp:Location (on body) |
| | pain | Sym:MSK:Pain + SymStatus:Experienced |
| Group 2 | Knee | SymProp:Location (on body) |

| | Pain | Sym:MSK:Pain + SymStatus:Experienced |
| --- | --- | --- |
| | never got any better | SymProp:Unchanged |
| | medication | SymRel:Alleviating Factor, No |

### Example 5 – Grouping Ambiguous Symptom Entities

| 55 | PT | "My stomach's feeling funny. It's been 3 days, but my stomach feels weird." |
|---|---|---|

*(30 lines down)*

| 85 | DR | "So, let's circle back to the last issue." |
|---|---|---|
| 86 | PT | "Yeah, my nausea. In the last 3 days, I've tried ginger tea and meds. Nothing helps." |

| Frame 1 | stomach's feeling funny | SymGI:Nausea + SymGI:Abdominal Pain + SymStatus:Experienced |
| --- | --- | --- |
| | 3 days | SymProp:Duration, All Time |
| | stomach feels weird | SymGI:Nausea + SymGI:Abdominal Pain + SymStatus:Experienced |
| Frame 2 | Nausea | SymGI:Nausea + SymStatus:Experienced |
| | ginger tea | SymRel:Alleviating Factor, Doesn't Help |
| | meds | SymRel:Alleviating Factor, Doesn't Help |

Takeaways:
- Although the patient is technically talking about the same problem when they mention "stomach's feeling funny" and much later "nausea", the initial mentions of this problem are ambiguous and have the same labels in the same order. Therefore, only those ambiguous mentions can be grouped together and the more clear (specific) mention is kept separate in a new frame.

# Important Reminders

## Condition as a Possible Symptom

Sometimes, a condition entity is mentioned that you might be able to interpret as a symptom. For example, the patient says, "The arthritis in my hand is getting really bad." You might want to interpret "arthritis" as Sym:MSK:Pain, and then "hand" as location, and "getting really bad" as severity. For this task, **we prefer that you DO NOT label any Condition entities as symptoms**. However, if you're not sure whether the condition mentioned is actually describing a symptom, you can label it, but be sure to tag with the SymSpecial:Unclear Condition label. Please note that the Unclear Condition tag should only be applied as an additional tag to Symptom Entities (**should not** be applied to attributes).

Example 6: SymSpecial:Unclear Condition

| 61 | PT | "My COPD seems to be worsening." |
| 62 | DR | "How so?" |
| 63 | PT | "I just feel like I can't do anything. I always have to use my oxygen." |
| 64 | DR | "And you're still smoking, right?" |

| Group 1 | COPD | Sym:Resp + SymStatus:Experienced + SymSpecial:Unclear Condition |
| | Worsening | SymProp:Worsening |
| | Can't do anything | SymProp:Severity |
| | oxygen | SymRel: Alleviating Factor, Unknown |

Takeaways:
- COPD is a condition. However, in this example the patient is describing it as something they experience, therefore it gets labeled.

# The Conditions Task

## Condition Entities

This task focuses on capturing every condition mentioned as entities, designating a status for each condition, and grouping with any additional related attributes.

**Condition** = objective classification of disease or physical disturbance/disorder denoted by the patient or provider; not to be mistaken with subjective experiences, which are categorized under Symptom.
- E.g., COPD is a condition, while coughing would be a symptomatic presentation of the condition.
    - **Exception**: If a patient is describing a chronic symptom (for our purposes, at least **6 months** or more), this should be captured as a condition.
        - E.g., "I've had bloating every day for years." "Bloating" would be considered a chronic symptom and should be captured under Conditions Task.
        - The patient or provider does not have to explicitly state the symptom has been occurring for more than six months. However, it should be clearly evident from the local context. If there is any doubt about the duration of the symptom, then we **do not** want to capture it as a condition.
- Reference this ICD-10 Code database if you need assistance determining if a span of text should be considered a condition or not. If it has an ICD-10 code, it is reasonable to label it as a condition.
- **Note:** Please use this as a resource only if you are having trouble deciding whether to highlight an entity or not. This **should not** be your first-line of information as there are numerous ICD-10 codes and many that overlap as symptoms based on our ontology. We are still relying on our labelers to use clinical judgement in determining what should be captured as a condition.
    - E.g., "Vomiting" is an ICD-10 code but that alone does not mean we would capture this as a condition.

### Condition Entity Labels + Applications

| | | |
|---|---|---|
| Condition | Patient | Applied when a condition is directly related to the patient. |

| Condition | Family History (Hereditary) | Applied when a heritable condition is mentioned in reference to a patient's family history of a blood-relative. Should NOT be used in reference to a patient's offspring.<br><br>Good Example:<br>**PT**: "My brother has high blood pressure."<br><br>Bad Example:<br>**PT**: "My son has asthma."<br>- While asthma is a heritable condition, it is not inheritable to the patient from their son. Therefore, the Family History (Hereditary) tag should **not** be applied. |
|---|---|---|
| Condition | Other | Applied to any "Other" mentions of conditions that aren't the patient's or hereditary family history. (i.e. a "catch-all" for conditions mentioned that don't fit the above descriptions, including education)<br><br>Examples:<br>**PT**: "My son has asthma." OR "My husband has diabetes." |

## Condition Attributes

The attributes we will be focusing on are: status, *time of onset/diagnosis, frequency, duration, improving, worsening, unchanged, location, severity, provoking factors, and alleviating factors*, as they relate to condition entities.

### Condition Attribute Labels + Applications

| ConditionStatus | Present | Applied to condition entities affirmed to be present for the **subject** in question.<br><br>Example:<br>**DR**: "Are you taking Lisinopril for your high blood pressure?"<br>**PT**: "Yes." |
|---|---|---|
| ConditionStatus | Absent | Applied to condition entities denied by the **subject** in question or for which no prior history has been communicated.<br><br>Example:<br>**DR**: "Any history of high blood pressure?"<br>**PT**: "No." |
| ConditionStatus | Unknown | Applied to condition entities for which we are unsure of the status for the **subject** or for which the **subject** is **at-risk.** |

| | | |
|---|---|---|
| | | Example:<br>DR: "Any history of high blood pressure or diabetes?"<br>PT: "I have diabetes."<br><br>Example:<br>DR: "Given your family history, you are at risk for heart disease." |
| ConditionStatus | Education | Applied to condition entities which are being discussed in an educational context.<br><br>Example:<br>DR: "High blood pressure is often asymptomatic." |
| ConditionAttr | Time of Onset/Diagnosis | When the subject was diagnosed with a particular condition.<br><br>Example:<br>PT: "Since I was 12." OR "A long time ago." |
| ConditionAttr | Frequency | Any specifications of the timing of how often a condition presents itself.<br><br>Example:<br>PT: "I get a sinus infection every 3 months." |
| ConditionAttr | Severity/Amount | Used to capture descriptive text regarding the **severity** OR **amount** of the condition being discussed.<br><br>Examples: "My anxiety is crippling OR I had 2 masses" |
| ConditionAttr | Duration | How long the subject has had a certain condition.<br><br>Example:<br>PT: "I've had psoriasis for 10 years now." |
| ConditionAttr | Duration, Flare-up | Used to capture text that indicates how long the subject has been experiencing a flare-up for a condition.<br><br>Example:<br>PT: "I've been going through a particularly bad bout of my psoriasis the past 3 weeks." |
| ConditionAttr | Characteristic/Quality | Used to capture any descriptive qualities related to a condition. Please **do not** capture symptoms as characteristics/quality for condition.<br><br>Example:<br>PT: "My migraines are typically throbbing."<br><br>Example:<br>PT: "Fortunately, your fracture is nondisplaced." |

| | | |
|---|---|---|
| | | Example:<br>PT: "Your asthma is what we call mild, intermittent."<br>- If the phrase "mild, intermittent asthma" is used, it is appropriate to capture the entire span as an entity. |
| ConditionAttr | Improving | Text indicating a condition is improving.<br><br>Examples: improving, better, not as bad<br><br>Use the surrounding context to determine the most appropriate label when deciding between improving/worsening/unchanged.<br><br>Example:<br>Pt: "Your diabetes is under control for the first time."<br>- You may be inclined to capture "under control" as unchanged, but the provider is indicating the diabetes has improved from past encounters based on the context. |
| ConditionAttr | Worsening | Text indicating a condition is worsening.<br><br>Examples: worsening, progressing, detoriarting<br><br>Use the surrounding context to determine the most appropriate label when deciding between improving/worsening/unchanged. |
| ConditionAttr | Unchanged | Used to capture text indicating a condition is unchanged.<br><br>Examples: about the same, unchanged, controlled, stable<br><br>Use the surrounding context to determine the most appropriate label when deciding between improving/worsening/unchanged. |
| ConditionAttr | Location | Specifications about where a condition presents itself on the body.<br><br>Example:<br>PT: My migraines are typically on the left side.<br><br>Example:<br>PT: I have compression fractures in my back. |
| ConditionAttr | Provoking Factor, Yes | Text indicating a cause for a condition worsening. Subject must confirm it worsens the condition.<br><br>For *Condition:Patient* entity, should **only** be applied if the |

| | | |
|---|---|---|
| | | patient has tried the provoking factor in question. Should **not** be used for questions around theoretical provoking factors.<br><br>Good Example:<br>**PT:** "==Chocolate== makes my migraines worse."<br><br>Good Example:<br>**DR**: "Does ==chocolate== make your migraines worse?"<br>**PT**: "Unfortunately, yes."<br><br>Bad Example:<br>**PT:** "So now I don't eat chocolate at all anymore."<br>- Even if the above statement is in the same conversation as the good examples from above, it would not be captured because in this context as it is not being discussed in a question or statement regarding its provocation.<br><br>Good Example:<br>**DR:** "Typically things like ==pollen== trigger people's allergies."<br>- When a provoking factor is discussed in an educational context, it should be captured appropriately. |
| ConditionAttr | Provoking Factor, No | Text indicating a potential cause for a condition worsening which has not worsened the condition in actuality. Subject must deny it worsens their condition.<br><br>For Condition:Patient entity, should **only** be applied if the patient has tried the provoking factor in question. Should **not** be used for questions around theoretical provoking factors.<br><br>Example:<br>**DR**: "Does ==chocolate== make your migraines worse?"<br>**PT**: "Thankfully, no." |
| ConditionAttr | Provoking Factor, Unknown | Used to capture concepts that may have worsened a condition, but the subject does not provide a clear explanation.<br><br>For Condition:Patient entity, should only be applied if the patient has tried the provoking factor in question. Should not be used for questions around theoretical provoking factors.<br><br>Good Example:<br>**DR: "**Has the recent ==stress== made your migraines any worse?"<br>**PT**: "Um, you know, I'm not really sure." |

| | | Bad Example:<br>**DR:** "Did you want to try Imitrex for your migraines?"<br>- Because Imitrex has not yet been tried by the patient, it would not be captured in this task. |
|---|---|---|
| ConditionAttr | Alleviating Factor, Yes | Text indicating a cause for a condition improving. Subject must confirm it improves their condition.<br><br>Example:<br>**DR:** "What are you taking for your depression?<br>**PT:** "I am taking Cymbalta. It's been helping." |
| ConditionAttr | Alleviating Factor, No | Text indicating a cause for a condition improving. Subject must deny it improves their condition.<br><br>Example:<br>**DR:** "What are you taking for your depression?<br>**PT:** "I am taking Cymbalta. It hasn't been helping." |
| ConditionAttr | Alleviating Factor, Unknown | Text indicating a potential cause for a condition improving, but not confirmed or denied by the patient.<br><br>For Condition:Patient entity, should **only** be applied if the patient has tried the alleviating factor in question. Should **not** be used for questions around theoretical alleviating factors.<br><br>Example:<br>**DR:** "What are you taking for your depression?<br>**PT:** "I am taking Cymbalta".<br>- The patient is taking the medication for their depression, but it is unclear whether the medication is helping the patient.<br><br>Example:<br>**DR**: "Did the Xifaxan help your IBS?"<br>**PT:** "Umm… I don't recall. I tried it a long time ago."<br>- The patient is taking the medication for their IBS, but they **do not** remember how effective it was in helping their condition. The medication may or may not have helped the patient's IBS. |
| ConditionSpecial | Resolved | Conditions that the subject previously had and has been confirmed to no longer have.<br><br>Note: This should **not** be applied to Family History or Other entities in which the subject is deceased. |
| ConditionSpecial | Irrelevant | Conditions that **do not** belong in the patient's medical record. |

### Example 1 - Grouping Entities

| | | |
|---|---|---|
| 92 | DR | "So, we're here to talk about your diabetes." |
| 93 | PT | "It was bound to happen. My mother and my father both have diabetes. " |
| 94 | DR | "What I have done with my other patients who have diabetes is start with lifestyle changes. Diabetes is one of the conditions that can see major improvements without medication if we focus on diet and exercise." |
| 95 | PT | "Let's start with that. I want to be able to try everything I can before starting a new medication." |

| Group 1 | diabetes (1st HL) | Condition: Patient + ConditionStatus: Present |
|---|---|---|
| Group 2 | My mother and my father both have diabetes (2nd HL) | Condition: Family History (Hereditary) + ConditionStatus: Present |
| Group 3 | diabetes (3rd HL)  diabetes (4th HL) | Condition: Other + ConditionStatus: Education  Condition: Other + ConditionStatus: Education |

Takeaways:
- The subjects "my mother and my father" are included in the highlighted span for heart disease because it better captures both concepts of Family History & Condition. When able, similar mentions of family conditions should be labeled this way.
- When there is a theoretical/education mention of a condition, the ConditionStatus:Education label should be applied.
- All entity tags must be given the appropriate status tag.

### Example 2 - Irrelevant Triple Tag

| | | |
|---|---|---|
| 50 | DR | Regarding your history of coronary artery disease, is there any family history of heart disease? |
| 51 | PT | Not that I know of [INAUDIBLE] I'm currently taking care of my stepson and he has diabetes. |

| Group 1 | coronary artery disease | Condition:Patient + ConditionStatus:Present |
|---|---|---|
| Group 2 | Family history of heart disease | Condition:Family History (Hereditary) + ConditionStatus:Unknown |
| Group 3 | my stepson and he has diabetes | Condition:Other + ConditionStatus:Present + ConditionSpecial:Irrelevant |

Takeaways:
- "Family history" is included in the highlight for heart disease because it better captures both concepts of Family & Condition. When able, similar mentions of family conditions should be labeled this way.
- The last mention of diabetes is given a status of Irrelevant, because it is not related to the patient's history nor their actual family history.
- The whole phrase "My stepson and he has" is included in the highlight for diabetes because it captures the subject matter that deems it irrelevant and allows us to standardize the capturing of subject matter for any subject other than the actual patient (which is specified by the actual Condition:Patient tag).

## Example 3 - Unchanged Attribute

111 DR "How long have you had asthma?"

112 PT "I was diagnosed around the age of 10."

113 DR "And how have you been doing? Any episodes where you can't seem to catch your breath?"

114 PT "Oh no, none of that. At least not anymore. It seems like everything is under control."

| Group 1 | Asthma | Condition:Patient + ConditionStatus:Present |
|---|---|---|
| | around the age of 10 | ConditionAttr:Time of Onset/Diagnosis |
| | Everything under control | ConditionAttr:Unchanged |

Takeaways:
- The attribute labels of worsening/improving/unchanged allow us to capture important information related to the entity (asthma) that would otherwise be lost.
- Note that "can't seem to catch your breath" is not captured. This is a symptomatic description of the patient's asthma, but not a description of the condition itself and therefore, we xb highlight under conditions task.

### Example 4 - Improving Attribute

| 34 | DR | "How about in your hands and stuff you have, well you have some arthritis there, does that bother you much?" |
| 35 | PT | "No, it doesn't bother me much." |
| 36 | DR | "That's great. Better than before?" |
| 37 | PT | "Oh yeah, much better! The pain used to be unbearable." |

| Group 1 | arthritis | Condition:Patient + ConditionStatus:Present |
| --- | --- | --- |
| | hands | ConditionAttr:Location |
| | doesn't bother me much | ConditionAttr:Severity |
| | much better | ConditionAttr:Improving |

Takeaways:
- We prioritize labeling the patient's interpretation of the condition when asked. If the DR were to make a statement about the status/experience, followed by the patient reiterating the same, we'd label both.
- The first mention of attribute, "doesn't bother me much" does not indicate if the condition is improving/worsening, but does describe the severity. Whereas, the second mention of status ("much better"), indicates the condition has improved.
- Note that we **do not** highlight "pain" or any of its related attributes such as "used to be unbearable" since this is more accurately classified as a symptom, not a condition. We only highlight actual conditions (and their attributes).

### Example 5 - Alleviating Factors

| 281 | DR | "We've talked about maybe putting you on a pancreas enzyme pill to see if that helps with the IBS. Would you be interested in doing that?" |

| Group 1 | IBS | Condition:Patient + ConditionStatus:Present |
| | pancreas enzyme pill | **No HL** |

Takeaways:
- Remember, for Condition:Patient we only capture provoking/alleviating factors that have been tried by the subject, not theoretical/future mentions. Based off of the context of this conversation, the patient has not yet tried taking a pancreas enzyme pill, therefore we **do not** capture this mention as an alleviating factor.

#### Example 6 - Condition as an Adjective

| 88 | DR | "Okay, so we will refill your diabetes medications and see you back in 3 months." |

| Group 1 | diabetes | **No HL** |

Takeaways:
- When a condition is mentioned as a descriptive term such as in the case of "diabetes medication," "colon cancer screening test" etc., we **do not** capture the condition as an entity. We only capture conditions as an entity when the condition itself is the subject rather than an adjective/modifier.

# The Treatment Task

## Treatment Entities

This task focuses on capturing treatments that are used to improve a condition, but not those used to prevent a condition from occurring.

**Treatment** = modality used to treat or improve the patient's condition.

- This ontology focuses on capturing surgeries and medical equipment.
    - **DO NOT** capture mentions of dialysis.
    - **DO NOT** capture mentions of procedural radiation.
    - **DO NOT** capture medications or non-med supplements.
    - **DO NOT** capture mentions of physical therapy or rehabilitation.
    - **DO NOT** capture non-medical therapies (acupuncture, chiropractic services, massage).

## Treatment Entity Labels + Applications

| | | |
|---|---|---|
| Treatment | Surgery | Used to capture mentions surgical procedures.<br><br>Only capture concrete mentions<br>● Avoid vague mentions that allude to the procedure.<br>● Avoid vague mentions that describe the procedure.<br><br>Only capture procedures that cure, mitigate, or treat.<br>● **Do not** capture procedures that diagnose or prevent (prophylactic procedures).<br><br>Stand-alone verbs should not be captured as a procedure, however, when paired with the appropriate noun/descriptor a verb may be appropriate to include in the entity span.<br>● Bad examples: replaced, taken out, removed<br>● Good Examples: hip replaced, appendix taken out<br><br>Good Examples:<br>● amputation<br>● cesarean section<br>● bariatric procedures<br>● transplant procedures<br>● genitourinary procedures<br>● gastrointestinal procedures<br>● cardiothoracic procedures<br>● neurosurgery procedures<br>● male reproductive health procedures<br>● female reproductive health procedures<br>● orthopedic procedures (replacement/repair)<br><br>Bad Examples:<br>● Biopsy<br>● Incision<br>● Cut it out<br>● Take it out<br>● Replaced<br>● Laparoscopic<br>● Vasectomy<br>● Tubal ligation |
| Treatment | Medical Equipment | Used to capture mentions of medical devices that are used for treatment<br><br>**Medical device** is defined as an apparatus, implant, instrument, machine (think nuts, bolts, screws). |

| | | Only capture devices that cure, mitigate, or treat.<br>● **Do not** capture devices that diagnosis or monitor.<br>● A glucose monitor or blood pressure machine are instrumental in a patient's care plan for each respective condition, but these devices are not used for treatment. We **do not** want to capture devices that aid in care.<br><br>Good Examples:<br>oxygen tank, crutches, wheelchair, insulin pump, glasses, wearable cardioverter defibrillator (WCD), ICD, CPAP machine, stent<br><br>Bad Examples:<br>bandages, gauze, compression socks, glucose monitor, blood pressure machine, needles, syringes, masks |
|---|---|---|

# Treatment Attributes

The attributes we will be focusing on are *status, tempo, location, response, and events*.

## Treatment Attribute Labels + Applications

| | | |
|---|---|---|
| TreatmentStatus | Present/Active | Applied to treatments the patient currently uses.<br><br>Applied to treatments the patient is scheduled to use.<br>- Tag may be used for future treatments that are scheduled or confirmed prior to the current encounter.<br><br>Example:<br>**PT**: "I use the CPAP machine every night."<br><br>Example:<br>**DR**: "Did you have the knee surgery?"<br>**PT**: "It is scheduled for next week." |
| TreatmentStatus | Absent/No HX | Applied to treatments the patient has never used or undergone.<br><br>Example:<br>**DR**: "Have you had knee surgery?"<br>**PT**: "No." |
| TreatmentStatus | Unknown | Applied to treatments where the status is unclear or not stated.<br><br>Applied to treatments that are mentioned in an educational or theoretical context.<br><br>Applied to treatments that **do not** belong in the patient's medical |

47

| | | |
|---|---|---|
| | | record (irrelevant mentions).<br><br>Example:<br>**DR**: "Have you had knee surgery prior?"<br>**PT**: "Well, I did have a procedure on my wrist."<br><br>Example:<br>**PT**: "I am here for a consultation for a fundoplication."<br><br>Example:<br>**PT**: "My sister had a hysterectomy." |
| TreatmentStatus | Past | Applied to treatments that have been completed in the past.<br><br>Example:<br>**DR**: "When was your pacemaker placed?"<br>**PT**: "2 years ago."<br><br>Example:<br>**DR**: "Have you had knee replacement surgery?"<br>**PT**: "Yes." |
| TreatmentAttr | Tempo | Used to capture **when** the patient receives a treatment (time of treatment).<br><br>Used to capture **how often** the patient receives a treatment (frequency).<br><br>Used to capture **how long** the patient receives a treatment (duration).<br><br>Example:<br>**PT**: "I use the CPAP machine every night."<br><br>Example:<br>**PT:** "I had my hysterectomy performed last year." |
| TreatmentAttr | Location | Used to capture the location of a treatment.<br><br>When a location precedes the entity, capture them as one highlight.<br>- Meaning **do not** use the location tag.<br><br>When a location succeeds (follows) the entity, capture them as two highlights.<br>- Meaning use the location tag.<br><br>Good Example:<br>**PT: "**The surgery was on my right knee." |

| | | |
|---|---|---|
| | | PT: "I had surgery on my ==left wrist.=="<br><br>Bad Examples:<br>PT: "I recently had back surgery."<br><br>PT: "I have my hip replacement surgery next month." |
| TreatmentAttr | Improving Worsening Unchanged (Response) | Used to capture spans that indicate the effects of treatment (improving/worsening/ unchanged).<br><br>Highlighted text will likely include a **symptom, condition, lab value.**<br><br>Highlighted text should capture the benefit or adverse effect itself.<br>- **Do not** capture phrases like better, improving, worsening<br><br>Good Example:<br>PT: I have had ==less pain== since my knee replacement.<br>- "less pain" is a benefit (beneficial response).<br>Good Example:<br>PT: I have had ==more pain== since my knee replacement.<br>- "more pain" is a side effect (adverse response .<br><br>Bad Example:<br>PT: The surgery seems to have helped.<br>- We want to capture the specific aspect of the patient's condition that has improved, not simply a statement that a treatment helped. |
| TreatmentEvent | Start | **Verb phrase** indicating a provider's direction or patient's acknowledgement to start using a treatment during the current encounter OR prior to the current encounter.<br><br>Patient does not have to start treatment immediately, as long as the plan to start a treatment is decided and agreed upon.<br><br>Only want to capture start events **initiated by a provider**.<br><br>Example:<br>DR: "You need to ==begin== using a portable oxygen tank."<br><br>Example:<br>PT: "You told me to ==start== using a portable oxygen tank months ago." |
| TreatmentEvent | Stop | **Verb phrase** indicating a provider's direction or patient's acknowledgement to stop using a treatment during the current encounter OR prior to the current encounter.<br><br>Patient does not have to stop treatment immediately, as long as the plan to stop treatment is decided and agreed upon. |

| | | Only want to capture stop events initiated by a provider.<br><br>Example:<br>**DR:** "Your leg seems to be healing well. You can stop using crutches."<br><br>Example:<br>**PT:** "Surgeon told me I could stop using my crutches a few weeks ago." |
|---|---|---|
| TreatmentEvent | Continue | **Verb phrase** describing a situation during the current encounter or prior to the current encounter where patient is instructed to continue using a treatment.<br><br>**Verb phrase** describing modifications made to an existing treatment during the current encounter or previous encounter.<br><br>Only want to capture Continue events **initiated by a provider**.<br><br>Examples: keep going, modify, tweak |

## Example 1 - Past Treatments

80    **DR**    "When did you have your meniscus repair surgery?"

81    **PT**    "Last month."

| Group 1 | Meniscus repair surgery | Treatment:Surgery<br>TreatmentStatus:Past |
|---|---|---|
| | Last month | TreatmentAttr:Tempo |

Takeaways:
- The meniscus repair has already occured, so we want to use "TreatmentStatus:Past"

## Example 2 - Duration and Frequency

92    **DR**    "So, how long have you been using CPAP?"

93    **PT**    "I have used it for two years now."

94    **DR**    "Ok and how often do you use it?"

95    **PT**    "Every night."

| CPAP | Treatment: MedicalEquipment<br>TreatmentStatus: Present/Active |
|---|---|
| two years | TreatmentAttr: Tempo |
| Every night | TreatmentAttr: Tempo |

Takeaways:
- Information related to time (time of onset, duration and frequency) will all be labeled using TreatmentAttr:Tempo.

### Example 3 - Concise Highlights

121    **DR**    Tell me about that surgery you had?

122    **PT**    Well it was about four months ago now.

| surgery | Treatment: Surgery<br>TreatmentStatus: Past |
|---|---|
| Four months ago | TreatmentAttr: Tempo |

Takeaways:
- When capturing an entity or attribute, the most concise span of text should be captured that accurately conveys the concept while excluding any information that may be irrelevant.
    - You may be inclined to capture "that surgery" or "about four months ago", but it is more concise to leave out "that" and "about".

---

# The Diagnostics Task

## Diagnostics Entities

This task focuses on capturing every diagnostic mention as an entity, designating a status for each diagnostic entity, and grouping with any related attributes.

**Diagnostic** = medical test used to screen for, monitor or diagnose a medical condition.

- **Do not** capture surgeries/procedures as these are captured in the treatment ontology.

## Diagnostics Entity Labels + Applications

| | | |
|---|---|---|
| Diagnostics | Labs/Pathology | Used to capture diagnostic laboratory/pathology studies.<br><br>**Do not** capture labs performed at home.<br><br>Examples: CBC, flu swab, strep test, blood glucose, labwork, blood test, hgbA1c, lipid panel, liver function tests, AST, ALT, metabolic panel, potassium, pregnancy test, pap smear, PSA, INR, TSH, urinalysis, urine culture, allergy testing |
| Diagnostics | Imaging | Used to capture diagnostic imaging studies.<br><br>Examples: CT, DEXA, Echo, mammogram, MRI, ultrasound, X-Ray |
| Diagnostics | Scopes | Used to capture scopes performed for diagnostic purposes.<br><br>Examples: colonoscopy, endoscopy, EGD |
| Diagnostics | Clinic Vitals | Used to capture mentions of clinic vital signs.<br><br>**Do not** capture vitals performed at home.<br><br>Examples: blood pressure, vitals, height, weight, pulse, oxygen, temperature, EKG/ECG, pulmonary function testing, incentive spirometry, stress test |
| Diagnostics | Home Labs/Vitals | Used to capture mentions of home labs or vitals.<br><br>**Do not** capture any labs/vitals performed in the clinic.<br><br>Examples: blood pressure, blood sugar, temperature, pulse, weight, height, oxygen saturation |

## Diagnostics Attributes

The attributes we will be focusing on are *status, improving, worsening, unchanged, location, indication, frequency, time of study, value/finding,* and *irrelevant*.

- We want entities to have clear associations with their attributes within frames. To eliminate ambiguity or confusion within frames, we ask that you **only capture the most recent attribute** when there are multiple mentions within an attribute category. **The only exception to this is the value-finding tags**, which are now differentiated by recent and prior findings. Example provided below.

## Diagnostics Attribute Labels + Applications

| | | |
|---|---|---|
| DiagnosticsStatus | Ordered | Applied to diagnostic entities that are pending from previous encounters or ordered during the current encounter.<br><br>**Do not** capture diagnostics that have been conducted or completed already.<br><br>Examples:<br>**DR:** "Here is a slip for your bloodwork."<br><br>**DR:** "I have sent the referral for your endoscopy." |
| DiagnosticsStatus | Completed | Applied to diagnostic entities that have been completed prior to the current encounter or during the current encounter.<br><br>Good Example:<br>**DR:** "Your blood pressure looks good today."<br><br>Bad Example:<br>**DR:** "Let's recheck your blood pressure before you leave."<br>- The repeat blood pressure has not been performed, so the completed tag should not be used. |
| DiagnosticsStatus | Requested | Applied to diagnostic entities requested by the patient.<br><br>Examples:<br>**PT:** "Can you recheck my PSA?"<br><br>**PT:** "It has been five years since my last mammogram. Can you order one today?" |
| DiagnosticsStatus | Not Performed | Applied to diagnostic entities that have not been performed or those that will not be performed.<br><br>Examples:<br>**DR:** "Have you had a colonoscopy in the past?"<br>**PT:** "No."<br><br>**DR:** "I understand you are claustrophobic. We will not do the MRI." |

| | | |
|---|---|---|
| DiagnosticsStatus | Unknown | Applied to diagnostic entities that have an unclear status for the patient or are irrelevant to the patient.<br><br>**Do not** capture diagnostics mentioned in an educational context.<br><br>Examples:<br>**PT:** "They either did a CT or MRI. I don't know which."<br><br>**DR:** "Did you have the DEXA scan?"<br>**PT:** "I think so." |
| DiagnosticsStatus | Education/Suggestion | Applied to diagnostic entities that are mentioned in an **educational or theoretical context** or diagnostics that are **suggested by the provider but not clearly ordered** during the current encounter.<br><br>Examples:<br>**DR:** "An echocardiogram is an ultrasound that looks at your heart."<br><br>**DR:** "Another option is an echocardiogram." |
| DiagnosticsAttr | Improving | Used for spans that indicate the diagnostic is improving.<br><br>Examples: improving, better, not as bad<br><br>Example:<br>**DR:** "Your A1C is better than we had in June." |
| DiagnosticsAttr | Worsening | Used for spans that indicate the diagnostic is worsening.<br><br>Examples: worsening, poorer, deteriorating<br><br>Example:<br>**DR:** "Your viral load is rapidly worsening." |
| DiagnosticsAttr | Unchanged | Used for spans that indicate the diagnostic is unchanged.<br><br>Examples: about the same, unchanged, stable<br><br>Example:<br>**DR:** "How is your blood pressure?"<br>**PT:** "It has been the same. Running around 150/90." |
| DiagnosticsAttr | Location | Used for spans that describe the location being evaluated by the diagnostic. |
| DiagnosticsAttr | Indication | Used for spans that describe why the diagnostic is being performed. |
| DiagnosticsAttr | Frequency | Used for spans that describe how often the diagnostic is |

| | | |
|---|---|---|
| | | performed or will be performed. |
| DiagnosticsAttr | Time of Study | Used for spans that describe when the diagnostic was conducted or will be conducted.<br><br>Note: Sample collection and testing can occur at different times for laboratory/pathology tests. We consider the time of collection to be the time of study in such instances. |
| DiagnosticsAttr | Value/Finding (Most Recent) | Applied to the **most recent** finding for that diagnostic.<br><br>It is appropriate to capture:<br>- Values (normal and abnormal).<br>- Results (normal and abnormal).<br>- Provider interpretation or commentary (mentions of improvement, deterioration, reassurance, non-reassuring commentary, etc).<br><br>Example:<br>**DR:** Your A1c is 6.0 now. That is great!<br>- We want to capture the numerical value.<br>- We also want to capture "great" because it is the doctor's interpretation of the result. |
| DiagnosticsAttr | Value/Finding (Prior) | Applied to previous findings for that diagnostic.<br><br>It is appropriate to capture…<br>- Values (normal and abnormal).<br>- Results (normal and abnormal).<br>- Provider interpretation or commentary (mentions of improvement, deterioration, reassurance, non-reassuring commentary, etc).<br><br>Example:<br>**DR:** Your A1c is 6.0 now. It was 6.5 in January earlier this year.<br>- "January" is associated with a previous lab value, so we would not capture it as time of study. |
| DiagnosticsSpecial | Irrelevant | Applied to diagnostics entities that are not relevant to the patient.<br><br>These spans would not be found in the patient's medical record.<br><br>Example:<br>**PT:** "My husband had a colonoscopy." |

## Example 1 - Grouping Entities

| | | |
|---|---|---|
| 92 | DR | "How have your sugars been running?" |
| 93 | PT | "Not much different, still around 200 or 220 in the mornings." |
| 94 | DR | "Yeah, it looks like your blood sugar was 232 when we checked it a few weeks ago." |
| 95 | PT | "Um-hum." |
| 96 | DR | "Let's check your A1c again and see." |

| Group 1 | sugars | Diagnostics:Home Labs/Vitals + DiagnosticsStatus:Completed |
|---|---|---|
| | Not much different | DiagnosticsAttr:Unchanged |
| | Around 200 or 220 | DiagnosticsAttr:Value/Finding (Most Recent) |
| | In the morning | DiagnosticsAttr:Time of Study |
| Group 2 | Blood sugar | Diagnostics:Labs/Pathology + DiagnosticsStatus:Completed |
| | 232 | DiagnosticsAttr:Value/Finding (Most Recent) |
| | A few weeks ago | DiagnosticsAttr:Time of Study |
| Group 3 | A1c | Diagnostics:Labs/Pathology + DiagnosticsStatus:Ordered |

Takeaways:
- Groups must contain spans with matching entity and status labels.
    - "blood glucose" is referencing a laboratory diagnostic.
    - "sugars" is referencing a home diagnostic (fingerstick glucose measurement).
    - These entities should be grouped separately as they are contextually different.

## Example 2 - Education/Suggestion Status

| | | |
|---|---|---|
| 41 | PT | "When do I have to get a colonoscopy?" |

| 42 | DR | "The recommendation is a screening colonoscopy beginning at age 50. So I will put in an order for you to get a colonoscopy. And then you're good for 10 years if it's negative." |

| | | |
|---|---|---|
| Group 1 | Colonoscopy (1st HL) | Diagnostics:Scope + DiagnosticsStatus:Unknown |
| Group 2 | Screening colonoscopy (2nd HL) | Diagnostics:Scope + DiagnosticsStatus:Education/Suggestion |
| | Beginning at age 50 | DiagnosticsAttr: Time of Study |
| | 10 years | DiagnosticsAttr:Frequency |
| Group 3 | Colonoscopy (3rd HL) | Diagnostic:Scope + DiagnosticsStatus:Ordered |

Takeaways:
- Colonoscopy (1st mention) does not have a clear status so "unknown" should be used.
- Colonoscopy (2nd mention) is educational regarding colorectal cancer screening guidelines so "education/suggestion" should be used.
- The attributes "beginning at age 50" and "10 years" are discussed in an educational context and should be grouped accordingly.
- We **do not** capture "10 years" in the same frame as the second mention of colonoscopy since we do not know if this is the true frequency of the diagnostic for the patient.
- Notice that the three mentions of colonoscopy are grouped separately because they are all discussed in different contexts and possess different status tags.

### Example 3 - Vital Signs

| 77 | DR | "It looks like you have gained some weight since your last visit. Up to 205 from 196." |
| 78 | PT | "Yeah." |
| 79 | DR | "What's changed? Are you watching what you eat?" |
| 80 | PT | "Not really. But I don't think that's really different." |

| 81 | DR | "122/78, that's good. It's just the weight we need to work on." |

| Group 1 | weight (line77) | Diagnostics:Clinic Vitals + DiagnosticsStatus:Completed |
|---|---|---|
| | Gained some (line 77) | DiagsAttr:Value/Finding (Most Recent) |
| | Up to 205 from 196 (line 77) | DiagnosticsAttr:Value/Finding (Most Recent) |

Takeaways:
- Changes in weight can be mentioned as a subjective symptom or objective finding. The patient's weight is being mentioned by the provider based on objective data from the visit. Given this context, weight should be captured using the "clinical vitals" entity tag.
- The value/finding tag is most appropriate for "gained some" because there is no way to ascertain whether this result is positive or negative.
- We **do not** capture "since your last visit" because it is too vague to be tagged as time of study.
- While clinical knowledge suggests that "122/78" and "good" are both referring to a blood pressure measurement, these should not be highlighted because there is no entity mentioned.
- We **do not** capture the second mention of weight because it is not stated as a diagnostic. This situation is similar to when elevated blood pressure measurements are discussed and the provider then states, "We need to get your blood pressure under control." later in the conversation. That second mention of blood pressure is referring to a condition and would not be captured in the diagnostics task.

# The Providers Task

## Provider Entities

This task focuses on capturing every relevant mention of a provider/specialty as an entity, designating a status for each provider/specialty, and grouping with related attributes or events.

**Provider** = registered healthcare professional OR specialty providing healthcare services.

- Provider entities include mentions of specialists or specialties that currently provide, have provided, or will provide healthcare to the patient.
    - Examples (provider): primary care doctor, cardiologist, psychiatrist, physician, physician assistant, nurse practitioner, nurse, physical therapist, chiropractor, eye doctor
        - Refer to ontology search terms for additional examples.
    - Examples (specialty): dermatology, neurology, emergency department
        - Refer to ontology search terms for additional examples.

## Provider Entity Labels + Applications

| | | |
|---|---|---|
| Provider | Current Encounter | Applied to the provider/specialty in the current encounter.<br><br>Span may be in regards to the present visit or follow-ups.<br><br>Appropriate to capture nouns like visit, appointment, etc.<br><br>Appropriate to capture spans like return to clinic, come see me, etc.<br><br>Example:<br>**DR**: "Schedule an appointment with my PA in 2 weeks."<br>- PA is referencing the clinic from the current encounter, so it is captured as Provider:Current Encounter.<br><br>Example:<br>**PT**: "So [PHYSICIAN NAME OTHER] when should I start this?" |
| Provider | Other | Applied to any provider/specialty outside the current encounter.<br><br>Examples: dermatologist, cardiologist, therapist, eye doctor, emergency department, hospital |

IMPORTANT NOTES:

- We **do not** want to capture pronouns.
    - Examples: he, she, me, I

- We want to capture mentions of provider types.
    - Examples: cardiologist, dermatologist

- We want to capture mentions of redacted PII.
    - Examples: [DEIDENTIFIED], [PHYSICIAN NAME OTHER]

- We want to capture phrases used in proxy for the provider or specialty.

- Examples: doctor, physician

# Provider Attributes

The attributes we will be focusing on are *status (present, absent, unknown), tempo, indication,* and *events (start, stop, and continue).*

## Provider Status/Attribute Labels + Applications

| | | |
|---|---|---|
| ProviderStatus | Present | Appropriate when the patient is established with a provider.<br><br>Should not be applied to providers the patient has not yet seen.<br><br>Example:<br>DR: "Let's have you back in 3 months".<br><br>Example:<br>DR: "You should follow up with your diabetes doctor for that." |
| ProviderStatus | Absent | Appropriate when the patient denies seeing a provider.<br><br>Appropriate when the patient is not established with a provider.<br><br>Example:<br>DR: "Do you see a cardiologist?"<br>PT: "No." |
| ProviderStatus | Unknown | Appropriate when a clear status is not communicated.<br><br>Appropriate when the patient is referred to a provider.<br><br>Appropriate when the patient is scheduled to see a provider (1st time).<br><br>Example:<br>DR: "Have you ever seen a cardiologist?"<br>PT: "A long time ago maybe."<br><br>Example:<br>DR: "Have you been to the chiropractor?"<br>PT: "I have an appointment for next week." |
| ProviderAttr | Tempo | Used for spans that describe **when** the patient will be seen (was seen).<br><br>Used for spans that describe **how often** the patient sees a provider.<br><br>Used for spans that describe **how long** the patient has seen a provider.<br><br>Example: |

| | | |
|---|---|---|
| | | DR: "I see my therapist ==next week==."<br><br>Example:<br>DR: "When did you last see your cardiologist?"<br>PT: "It was probably ==almost a year ago==." |
| ProviderAttr | Indication | Used for spans that describe why the patient is being seen or referred.<br><br>Indications can be symptoms, conditions, treatments, etc.<br><br>Examples: cancer, radiation, diabetes, hip surgery |
| ProviderEvent | Start | **Verb phrase** indicating the provider's direction or patient's acknowledgement to start seeing a provider<br><br>Example:<br>DR: "Let's ==start== you with Physical Therapy for that."<br><br>Example:<br>PT: "You ==referred== me to Neurology but I haven't been able to go."<br><br>For start events initiated prior to the current encounter, apply the double tag "ProviderSpecial:Past".<br><br>Example:<br>PT: "Last time you ==referred== me to Neurology but I haven't been able to get an appointment." |
| ProviderEvent | Stop | **Verb phrase** indicating the provider's direction or patient's acknowledgement to stop seeing a provider.<br><br>Should not be used if the patient stopped seeing the provider without the direction or agreement of another provider.<br><br>Good Example:<br>DR: "You can ==stop seeing== the chiropractor since you do not feel better."<br>Bad Example:<br>PT: "I stopped seeing the chiropractor."<br>- It seems the patient made the decision to stop seeing the provider out of their own volition. Since this action is not clearly directed by a provider, we would not capture this span as a stop event.<br><br>For stop events initiated prior to the current encounter, apply the double tag "ProviderSpecial:Past". |

| | | |
|---|---|---|
| | | Example:<br>PT: "The stomach doctor told me I could stop coming to them unless I had any other concerns." |
| ProviderEvent | Continue | **Verb phrase** indicating the patient should continue to follow up with a current provider.<br><br>Good Example:<br>DR: You can keep going with the chiropractor if it helps.<br><br>Bad Example:<br>PT: "I have decided to keep going to the chiropractor."<br>- It seems the patient made the decision to stop seeing the provider out of their own volition. Since this action is not clearly directed by a provider, we would not capture this span as a continuation event.<br><br>For continuation events initiated prior to the current encounter, apply the double tag "ProviderSpecial:Past".<br><br>Example:<br>PT: "Last time you told me you wanted me to keep coming every 3 months." |
| ProviderSpecial | Past | Applied as a triple tag when the patient is no longer seeing a provider<br><br>Applied as a triple tag when events (start, stop, continue) were initiated in the past<br><br>Example:<br>PT: "I used to see a dermatologist." |

## Example 1 - Grouping

92    **DR**    "Let's refer you to Urology. Also, are you still following up with your allergist?"

93    **PT**    "Ok, that sounds good. Yes, I do once a month."

| Group 1 | refer | ProviderEvent:Start |
|---|---|---|
| | urology | Provider:Other<br>ProviderStatus:Unknown |

|  |  |  |
|---|---|---|
| Group 2 | Allergist | Provider:Other<br>ProviderStatus:Present |
|  | once a month | ProviderAttr:Tempo |

Takeaways:
- Remember to separate each provider entity into different frames if they are not conceptually synonymous.

## Example 2 - Educational Mentions or Suggestions

95  **DR**  "I usually recommend patients with similar symptoms to see a neurologist."

96  **PT**  "Okay."

97  **DR**  " I am putting in the referral to the neurology office."

|  |  |  |
|---|---|---|
| Group 1 | putting in the referral | ProviderEvent:Start |
|  | Neurology office | Provider:Other<br>ProviderStatus:Unknown |

Takeaways:
- We **do not** want to capture "neurologist" because it is an educational mention.
    - We **do not** want to capture educational mentions in this task.

## Example 3 - Event, Current Encounter

*NOTE: Encounter type of the example below is OB-GYN.

98  **DR**  "It's important for you to follow up with OB. Let's have you back in one year."

99  **PT**  "Ok, thanks."

|  |  |  |
|---|---|---|
| Group 1 | follow up | ProviderEvent:Continue |
|  | OB | Provider:Current Encounter<br>ProviderStatus:Present |
|  | have you back | ProviderEvent:Continue |

|  | one year | ProviderAttr:Tempo |
|---|---|---|

Takeaways:
- Be mindful of the current encounter type listed on the upper-right hand corner of the conversation. It is possible to have mentions of the current provider's specialty within the conversation. If this occurs, apply "Provider:Current Encounter".
- Remember to group "Provider:Current Encounter" with relevant attributes regarding the patient's next follow up with the current provider.

### Example 4 - Event, Referrals

41    **DR**    "I referred you to the pain specialists for your back pain last year. Did you see them?"

42    **PT**    "Um-hum."

43    **DR**    "Okay, when was the last time you saw him?"

44    **PT**    "[DEIDENTIFIED] has me come in every 3 months. I have an appointment in 2 weeks."

| Group 1 | referred | ProviderEvent:Start<br>ProviderSpecial:Past |
|---|---|---|
|  | pain specialists | Provider:Other<br>ProviderStatus:Present |
|  | back pain | ProviderAttr:Indication |
|  | [DEIDENTIFIED] | Provider:Other<br>ProviderStatus:Present |
|  | every 3 months | ProviderAttr:Tempo |
|  | appointment | Provider:Other<br>ProviderStatus:Present |
|  | 2 weeks | ProviderAttr:Tempo |

Takeaways:
- "last year" is not captured because it refers to when the provider placed the referral and not necessarily when the patient saw the pain specialists.

- The provider states they referred the patient in the past, so "referred" is captured as "ProviderEvent:Start" and "ProviderSpecial:Past".
- Please note that PII spans of text (e.g. [DEIDENTIFIED]) are captured when it is clear the span of text is referring to a medical provider or specialty.

### Example 5 - Capturing PII as Provider Entity

| 45 | DR | "You did your radiation over with [PHYSICIAN NAME OTHER], correct?" |
| 46 | PT | "Mmhmm." |
| 47 | DR | "Okay, good. And when did he last see you?" |
| 48 | PT | "It's been awhile, but I see him next month." |

| Group 1 | Radiation | ProviderAttr:Indication |
| --- | --- | --- |
| | [PHYSICIAN NAME OTHER] | Provider:Other<br>ProviderStatus:Present |
| | It's been awhile | ProviderAttr:Tempo<br>ProviderSpecial:Past |
| | Next month | ProviderAttr:Tempo |

Takeaways:
- Although pronouns are clearly being used to refer to a provider/specialty, we do not capture pronouns because it does not provide the modeling team with a consistent reference.
- ProviderAttr:Indication can be used to capture any span that explains why the provider/specialty is being seen.

# The Social History Task

## Social History Entities

This task focuses on capturing social history entities, designating a status for each entity, and grouping with any related attributes.

**Social History** = social factors that are relevant to the medical chart.
- <u>Relevant</u> social history entities…
    - Impact the patient's physical or mental health.
    - Impact the provider's medical decision making or the clinical course.

## Social History Entity Labels + Applications

| Social History | Alcohol Use | Used to capture mentions of alcohol use (active or inactive).<br><br>Should only be applied to general spans like alcohol, drink, etc.<br><br>Specific details regarding the patient's alcohol consumption should be captured using the "Characteristic/Details" tag.<br><br><u>Good Example:</u><br>**DR**: "Any alcohol use?"<br>**PT**: "No."<br><br><u>Bad Example:</u><br>**PT**: "Mostly beer a few times a week."<br>- This span describes the patient's alcohol consumption and should be tagged with "characteristic/details". |
|---|---|---|
| Social History | Children | Used to capture mentions of children (having or raising).<br><br>**Do not** capture pronouns.<br><br>**Do not** capture additional family members.<br><br><u>Good Examples:</u> daughter, son, raising my nephew |
| Social History | Diet | Used to capture dietary habits, changes, and programs.<br><br>Should only be applied to general spans like keto diet, etc.<br><br>Specific details regarding the patient's dietary habits should be captured using the "Characteristic/Details" tag.<br><br>**Do not** capture dietary supplements.<br><br><u>Good Examples:</u> keto diet, mediterranean diet, well-balanced diet, eating better, eating more vegetables<br><br><u>Good Examples:</u><br>**PT**: "I cut back on carbs."<br>- This is an overall change in diet being reported by the |

| | | patient and should be captured.<br><br>Bad Examples:<br>PT: "I take apples to work."<br>PT: "I eat less ice cream."<br>PT: "I don't eat chips."<br>- These are single items the patient is avoiding in their diet and should not be captured as entities. |
|---|---|---|
| Social History | Employment | Used to capture mentions of the patient's employment status or job information.<br><br>Good Example:<br>PT: "I ==retired== a few years ago."<br><br>Good Example:<br>DR: "What do you do for work?"<br>PT: "I ==mostly work at a desk==."<br><br>Bad Example:<br>PT: "My son is a lawyer."<br>- This information is not relevant to the patient's health or the provider's clinical decision making, so we would not capture it. |
| Social History | Illicit/Recreational Drug Use | Used to capture mentions of illicit drug use.<br><br>This does not include prescription medications that are abused by the patient (captured in medications task).<br><br>Examples: marijuana, cocaine, meth, heroin |
| Social History | Living Condition | Used to capture mentions of housing status.<br><br>Examples: homeless, housed in a shelter, bought a house |
| Social History | Marital Status | Used to capture the patient's marital status.<br><br>Examples: married, single, divorced, domestic partner, wife, husband |
| Social History | Physical Activity | Used to capture physical activity other than physical therapy exercises.<br><br>Should **only** be applied to general spans like exercise, activity, etc. |

|  |  | Specific details regarding the patient's physical activity should be captured using the "Characteristic/Details" tag. |
|---|---|---|
|  |  | Good Example:<br>PT: "I try to exercise at least 3 days a week."<br><br>Good Example:<br>DR: "Are you active?"<br>PT: "Not really."<br><br>Bad Example:<br>PT: "I have been running more lately."<br>- "running" describes how the patient exercises and should be captured using "characteristic/details".<br>- This should only be highlighted if an associated entity like "exercise" is available. |
| Social History | Sexually Active | Used to capture mentions of sexual activity. |
| Social History | Tobacco Use | Used to capture mentions of tobacco use.<br><br>Examples: smoking, chewing tobacco, cigarettes, cigars |
| Social History | Other | "Catch all" for social history that is relevant to the patient's health or clinical course but not otherwise applicable to the defined categories.<br><br>Good Examples: ADLs, environmental exposures, religious affiliation<br><br>**Do not** capture additional mentions of family members unless paired with other relevant social history information<br><br>Good Examples: my dad lives with us, my mom is in the hospital<br><br>Bad Examples: father, mother, brother, cousin |

# Social History Attributes

The attributes we will be focusing on are *status, tempo, progression, characteristic/detail,* and *modifying factors.*

## Social History Status/Attribute Labels + Applications

| SocialStatus | Present | Applied to social history explicitly stated as true or present. |
|---|---|---|

| | | |
|---|---|---|
| SocialStatus | Absent | Applied to social history explicitly stated as false or absent. |
| SocialStatus | Unknown | Applied to social history where status is unclear or not stated.<br><br>**Do not** use for theoretical mentions. |
| SocialAttr | Tempo | Used to capture **when**, **how often**, or **how long** a patient partakes in a given piece of social history.<br><br>Good Example:<br>**PT:** "I started smoking during high school, quit for awhile, then started again a few years ago."<br>- These spans describe onset and should be captured.<br><br>Good Example:<br>**PT:** "I have been smoking for several years."<br>- This span describes duration and should be captured.<br><br>Examples: on and off, regularly, days, weeks, months, years |
| SocialAttr | Progression | Used to capture text indicating if an element of social history is improving, worsening, or unchanged.<br><br>Good Example:<br>**DR**: "How's the alcohol use?"<br>**PT**: "Doing better, down to 3 drinks a day."<br>- These spans are describing an improvement in patient's alcohol use based on the context.<br><br>Examples (improving): improving, recovering<br><br>Examples (worsening): worsening, declining<br><br>Examples (unchanged): about the same, controlled |
| SocialAttr | Characteristic/Details | Used to describe characteristics of the entity.<br><br>Example:<br>**DR**: "Are you active?"<br>**PT**: "I try to be. I run a few times a week."<br>- Specifics of how the patient is active (entity) are captured using this tag.<br><br>Example: |

69

| | | PT: "I cut back on carbs. I'm eating less bread and chips."<br>- Specifics of how the patient cut back on carbs (entity) are captured using this tag.<br><br>Example:<br>DR: "Any alcohol use?"<br>PT: "Sometimes, mostly wine."<br>- Specifics of the patient's alcohol use (entity) are captured using this tag.<br><br>Example:<br>PT: "I smoke about 2 packs everyday."<br>- The number of packs the patient smokes are captured using this tag. |
|---|---|---|
| SocialAttr | Modifying Factor, Yes | Used to capture confirmed provoking factors and alleviating factors for social behaviors.<br><br>Example:<br>DR: "Have you been drinking more?"<br>PT: "Yeah. There has been a lot of stress at work." |
| SocialAttr | Modifying Factor, No | Used to capture factors that **do not** provoke or alleviate social behaviors. |
| SocialAttr | Modifying Factor, Unknown | Used to capture factors that are not clearly stated to provoke/alleviate or not provoke/not alleviate a particular social behavior.<br><br>Used when alleviation or provocation is not explicitly stated.<br><br>**Should not** be used to capture theoretical provoking factors that have not been tried by the patient. |
| SocialSpecial | Resolved | Applied to social history that is no longer true for the patient (has cleared or healed).<br><br>Example:<br>DR: "Are you still drinking?"<br>PT: "Not anymore. I stopped months ago."<br>- Since the patient no longer drinks, we would triple tag this span with "resolved". |

## Example 1 - Social History Other

45    **MA**    "Anything else changed in your medical history? New surgeries? Any changes to

your family history?"

46     **PT**     "This is not really medical, but I just moved to the area a couple of weeks ago, from the east coast."

| Group 1 | moved to the area | Social History:Other<br>SocialStatus:Present |
|---|---|---|
| | a couple of weeks ago | SocialAttr: Tempo |
| | from the east coast | SocialAttr:Characteristic/Details |

Takeaways:
- "moved to the area" is important social history to capture. It does not fit clearly into any of the other social history entity categories, so we choose the "other" entity tag.
- "a couple of weeks ago" describes when the patient moved and how long he/she has been in the area, so we label that span as "SocialAttr:Tempo".
- "from the east coast" describes the patient's move, so we label that span as "SocialAttr: Characteristic/Details".

## Example 2 - Social History Attributes

126     **DR**     "Do you smoke at all?"

127     **PT**     "No, I stopped when I was 25."

128     **DR**     "Oh great. About how many years did you smoke?"

129     **PT**     "3 or 4 years."

130     **DR**     "Okay. Well I'm very glad you quit. Any alcohol?"

131     **PT**     "Yes, sometimes. I did binge this past weekend though. I went to a wedding."

132     **DR**     "Oh well, let's keep it to 1 to 2 if we can."

| Group 1 | Smoke | Social History:Tobacco Use<br>SocialStatus:Present<br>SocialSpecial:Resolved |
|---|---|---|
| | stopped when I was 25 | SocialAttr:Tempo |

|  | Smoke | Social History:Tobacco Use SocialStatus:Present Social Special:Resolved |
|---|---|---|
|  | 3 or 4 years | SocialAttr:Tempo |
| Group 2 | alcohol | Social History:Alcohol Use SocialStatus:Present |
|  | sometimes | SocialAttr:Tempo |

Takeaways:
- Although the patient is no longer an active smoker, this is still pertinent to their medical record. In order to indicate this, we assign a status of "present" and add the triple tag of "resolved".
- It is important to capture how long the patient smoked for and when they quit. Both these spans should be captured with "SocialAttr:Tempo".

## Example 3 - Physical Activity

| 43 | DR | "How about exercise? Do you walk or have a routine?" |
| 44 | PT | "Well, there are stairs in my house. Does that count?" |

| Group 1 | exercise | Social History:Physical Activity SocialStatus:Present |
|---|---|---|
|  | stairs in my house | SocialAttr:Characteristic/Details |

Takeaways:
- Although the patient does not directly acknowledge a formal exercise routine, they do refer to some aspect of physical activity. Whether or not the physician counts that as exercise is not a call that we need to make.
- "Stairs in my house" is a descriptor of the physical activity the patient does, and should be labeled as a "Characteristic/Detail".



\end{document}